\documentclass[10pt,twocolumn,letterpaper]{article}

\usepackage[algorithms]{wacv}  
\usepackage{xcolor}
\usepackage[accsupp]{axessibility}  

\usepackage{pifont}
\newcommand{\tick}{\ding{51}}
\newcommand{\cross}{\ding{55}}

\usepackage{pifont}

\usepackage{graphicx}
\usepackage{subcaption}
\usepackage{placeins}
\usepackage{dblfloatfix}
\usepackage{multirow} 
\usepackage{booktabs}
\usepackage{xcolor}
\usepackage{tabularx}
\usepackage[normalem]{ulem}
\usepackage{makecell}
\usepackage{comment}
\usepackage{tabularx}
\usepackage{placeins}
\usepackage{tikz}
\usepackage{float}
\usepackage{booktabs}
\usepackage{arydshln}

\definecolor{wacvblue}{rgb}{0.21,0.49,0.74}
\usepackage[pagebackref,breaklinks,colorlinks,allcolors=wacvblue]{hyperref}
\usepackage{adjustbox}

\def\wacvPaperID{xxx} 
\def\confName{WACV}
\def\confYear{2027}

\title{Geometry-Aware Trajectory State Modeling for Few-Shot Action Recognition}

\author{
\begin{adjustbox}{max width=\textwidth}
Zhiqi Miao$^{1}$\quad
Marko Haralović$^{2,3}$\quad
Stanislav Levendeev$^{2}$\quad
Sarvesh Angadi$^{2}$
\end{adjustbox}\\
\begin{adjustbox}{max width=\textwidth}
Jannatul Ferdous$^{2}$\quad
Dan-Cristian Ploesteanu$^{2}$\quad
Jiapan Guo$^{1}$\quad
Estefanía Talavera$^{2}$
\end{adjustbox}\\[2pt]
\begin{adjustbox}{max width=\textwidth}
$^{1}$University of Groningen\enspace
$^{2}$University of Twente
\end{adjustbox}\\
\begin{adjustbox}{max width=\textwidth}
$^{3}$Faculty of Electrical Engineering and Computing, University of Zagreb
\end{adjustbox}\\[2pt]
{\ttfamily\small z.miao@rug.nl,\ e.talaveramartinez@utwente.nl}
}

\begin{document}
\maketitle
\begin{abstract}
 Few-shot action recognition (FSAR) requires representations that expose discriminative motion dynamics from only a few labeled videos. Recent trajectory-based methods leverage long-term point tracks for explicit temporal correspondence, but often treat trajectories as carriers of sampled visual features or summarize their motion with compact 2D displacement-based descriptors. We instead view an action as the ordered evolution of 2.5D trajectory states and their local geometric interactions. Based on this view, we introduce \textbf{GeoTRAM}, a geometry-aware trajectory state modeling framework for FSAR. GeoTRAM lifts sparse tracks into 2.5D trajectories, where \textbf{StateMotion} models ordered intra-trajectory state evolution and \textbf{RelMotion} captures state-conditioned local interactions through time-varying relative geometry and motion. Quality-filtered spectral sampling provides more reliable trajectory initialization; complementary semantic, motion, and depth cues are separately refined and preserved for joint spatio-temporal modeling; action-guided reweighting and top-\(\rho\) trajectory selection support complementary frame- and trajectory-level matching. Experiments on five FSAR benchmarks demonstrate strong gains on the motion-centric Something-Something benchmarks while maintaining competitive performance on more appearance-driven benchmarks and a favorable accuracy--efficiency trade-off. Our code is publicly available.\footnote{Code: \url{https://github.com/MiaoMeow98/GeoTRAM}}
    
\end{abstract}
\section{Introduction}
\label{sec:intro}

Recognizing an action requires more than identifying what appears in a video; it requires understanding how objects and body parts move, interact, and evolve over time. This distinction becomes particularly important in few-shot action recognition (FSAR), where a model needs to recognize novel action categories from only a few labeled videos. Existing FSAR methods have made substantial progress by comparing support and query videos through temporal alignment, set matching, or learned similarity functions~\cite{cao2020otam,zhang2020few,perrett2021trx,wang2022hyrsm,wang2023molo,lee2025team}. However, limited supervision makes it difficult to infer discriminative motion patterns purely from appearance representations. Consequently, an effective FSAR representation should retain semantic appearance while exposing temporal dynamics in a form that can be reliably compared from few examples.

Recent progress in long-term point tracking provides a promising paradigm for this purpose~\cite{doersch2023tapir,karaev2024cotracker,karaev2025cotracker3,xiao2024spatialtracker}. Unlike short-term frame differences or pairwise optical flow, tracked points maintain explicit correspondences over time and provide a compact way to describe how local image regions move through a video. Building on this property, trajectory-based FSAR methods~\cite{kumar2024tats,kumar2025trokens} associate tracked points with self-supervised visual features~\cite{oquab2024dinov2}, showing that sparse trajectories can provide useful motion-aware evidence for few-shot recognition. However, the availability of reliable tracks raises a more fundamental representation question: \textbf{what information should a trajectory encode in order to describe an action?}

We argue that a trajectory should not be treated only as a carrier of sampled visual features or summarized only with compact 2D displacement-based motion descriptors. An action is inherently a temporally ordered dynamical process. The same overall displacement patterns may arise from different state evolutions, while reversing or altering the order of local motion can correspond to a different action. What matters is therefore not only where a point moves, but also how its motion state evolves over time and how this evolution relates to nearby trajectories. This motivates a trajectory-state view of action representation: each trajectory is described as an ordered sequence of explicit kinematic states, and action dynamics are modeled through both the evolution of individual states and their local interactions.

This view naturally leads to two complementary levels of trajectory dynamics. At the intra-trajectory level, an action should preserve how the motion state of each tracked point evolves over time, rather than representing trajectory dynamics primarily through displacement-derived descriptors. We first augment image-plane coordinates with relative monocular depth to construct the 2.5D trajectory, exposing depth-axis variation that is unavailable in 2D. At each time step, we then explicitly describe its motion state using interpretable attributes including position, velocity, acceleration, depth variation, and visibility, and preserve the ordered state sequence for lightweight temporal encoding. At the inter-trajectory level, many actions are characterized by local interactions among moving parts or objects. Rather than modeling all trajectory pairs, we restrict interaction modeling to spatially nearby trajectories and describe their time-varying relative geometry and motion. Because proximity alone does not imply equal relevance, neighbor relations are further weighted by attention conditioned on both the current intra-trajectory state and their pairwise relation. In this way, geometric proximity determines which interactions are plausible, while state-conditioned attention determines which interactions are relevant. Together, these two levels capture both the temporal evolution of individual trajectories and their evolving local interactions.

Effective trajectory-state modeling further depends on \textbf{where trajectories originate}, \textbf{how heterogeneous evidence enters spatio- temporal modeling}, and \textbf{ which trajectories ultimately support recognition}. First, uniform or semantic-aware sampling may allocate a limited tracking budget to background, fragmented, or temporally unstable regions, weakening the evidence extracted along the resulting tracks. Second, semantic, depth, and motion cues encode complementary but heterogeneous information, and prematurely collapsing them through element-wise addition may obscure cue-specific information for joint spatio-temporal modeling. Finally, even reliably tracked trajectories may not be equally informative for the target action. This motivates three complementary designs: combining spectral clustering, quality filtering, and budgeted seed allocation to improve trajectory initialization; refining cue-specific information in dedicated subspaces followed by concatenation to preserve heterogeneous cues for joint spatio--temporal modeling; and leveraging class-token guidance to prioritize action-relevant trajectories.

Based on these principles, we propose \textbf{GeoTRAM}, a geometry-aware trajectory state modeling framework for few-shot action recognition. GeoTRAM first obtains reliable tracking seeds through quality-filtered spectral sampling and augments the resulting image-plane tracks with relative monocular depth to construct 2.5D trajectories. It then models their dynamics at two complementary levels: \textbf{StateMotion} encodes the ordered evolution of explicit intra-trajectory kinematic states, while \textbf{RelMotion} captures state-conditioned local interactions through time-varying relative geometry and motion. The resulting semantic, depth, and motion cues are temporally refined in dedicated subspaces, concatenated, and jointly processed by a divided space--time Transformer. Finally, DINOv2~\cite{oquab2024dinov2} semantic and video-level class tokens jointly guide trajectory relevance estimation and reweighting, with top-$\rho$ trajectory selection supporting trajectory-level instance matching alongside complementary frame-level prototype matching. Extensive experiments on five FSAR benchmarks demonstrate strong performance of GeoTRAM on the motion-centric Something-Something benchmarks while remaining competitive on more appearance-driven benchmarks, with a favorable accuracy--efficiency trade-off.

\section{Related Work}
\label{sec:related_work}

\noindent\textbf{Few-Shot Action Recognition.}
Few-shot action recognition (FSAR) aims to recognize novel action categories from only a few labeled videos. Most methods adopt episodic metric learning, where support and query videos are encoded into a shared representation space and compared therein. One line of work focuses on more effective support--query matching through temporal alignment, set matching, relation modeling, alignment-free comparison, and prototype learning~\cite{cao2020otam,zhang2020few,perrett2021trx,wang2022hyrsm,wang2024hyrsm++,zhang2021ITANet,thatipelli2022strm,lee2025team,yang2026MPL}. Another line strengthens video representations through richer temporal reasoning, explicit motion modeling, and multimodal cues~\cite{zhu2018compound,Fu2020DepthMetaFusion,xing2023gghm,wanyan2023amfar,xing2023SloshNet,wang2023molo, Liu_2023_WACV, tang2024safsar}. 
More recent studies further explore parameter-efficient adaptation, state-space architectures, and language- or prompt-guided priors~\cite{pei2025DS2Adapter,huang2025manta,qian2025LGA,qi2026P2A,Ponbagavathi_2026_Frame2Freq}. In contrast, our work focuses on the more specific question of how a sparse trajectory itself should encode action dynamics in the few-shot scenario.

\noindent\textbf{Trajectory Representations for Action Understanding.} Motion trajectories have long provided explicit structure for action understanding. Classical methods characterize tracked motion through trajectory shape, displacement statistics, optical flow descriptors, and local spatio-temporal features~\cite{laptev2008learning,wang2013action,wang2015action,pervs2010histograms}. These methods often summarize motion using compact displacement- or histogram-based descriptors~\cite{gowayyed2013hod}. More recent approaches leverage 3D human pose and tracklets for action recognition~\cite{rajasegaran2023benefits}, model absolute and relative body-point motion~\cite{huang2025hmore}, or organize video representations into temporally coherent trajectory tokens~\cite{zheng2025trajvit,zheng2026trajtok}. Meanwhile, modern point trackers enable reliable long-range correspondences~\cite{doersch2023tapir,karaev2024cotracker,karaev2025cotracker3,zholus2025tapnext,locotrack}, while 3D-aware tracking extends trajectories beyond the image plane~\cite{xiao2024spatialtracker,xiao2025spatialtrackerv2}. These advances improve the reliability and geometric richness of tracked trajectories. GeoTRAM focuses on how sparse trajectories can encode action dynamics for FSAR, explicitly representing each as an ordered 2.5D motion-state sequence and modeling time-varying, state conditioned local interactions among neighboring trajectories.

\noindent\textbf{Trajectory-Based FSAR.}
Trajectory-based FSAR has recently emerged as a promising paradigm for exposing long-term motion under limited supervision. TATs~\cite{kumar2024tats} uses point trajectories~\cite{karaev2024cotracker} as temporal anchors for aligning self-supervised visual features~\cite{oquab2024dinov2} and constructs trajectory-aligned representations that are subsequently processed by spatio-temporal encoding. Trokens~\cite{kumar2025trokens} further introduces semantic-aware trajectory initialization together with explicit intra- and inter-trajectory motion descriptors, including histogram-based modeling of intra-trajectory displacement and full relational modeling between trajectories. These methods establish sparse trajectories as effective motion carriers for FSAR, but trajectory dynamics are still represented either implicitly through aligned visual features or explicitly through histogram-based 2D displacement descriptors and all-pair inter-trajectory relations. GeoTRAM instead adopts a trajectory-state perspective, representing each relative 2.5D trajectory as an ordered sequence of explicit motion states while modeling local interactions through state-conditioned neighbor relations, with other components together supporting the GeoTRAM.

\begin{figure*}[!h]
    \centering
    \includegraphics[width=0.84\linewidth]{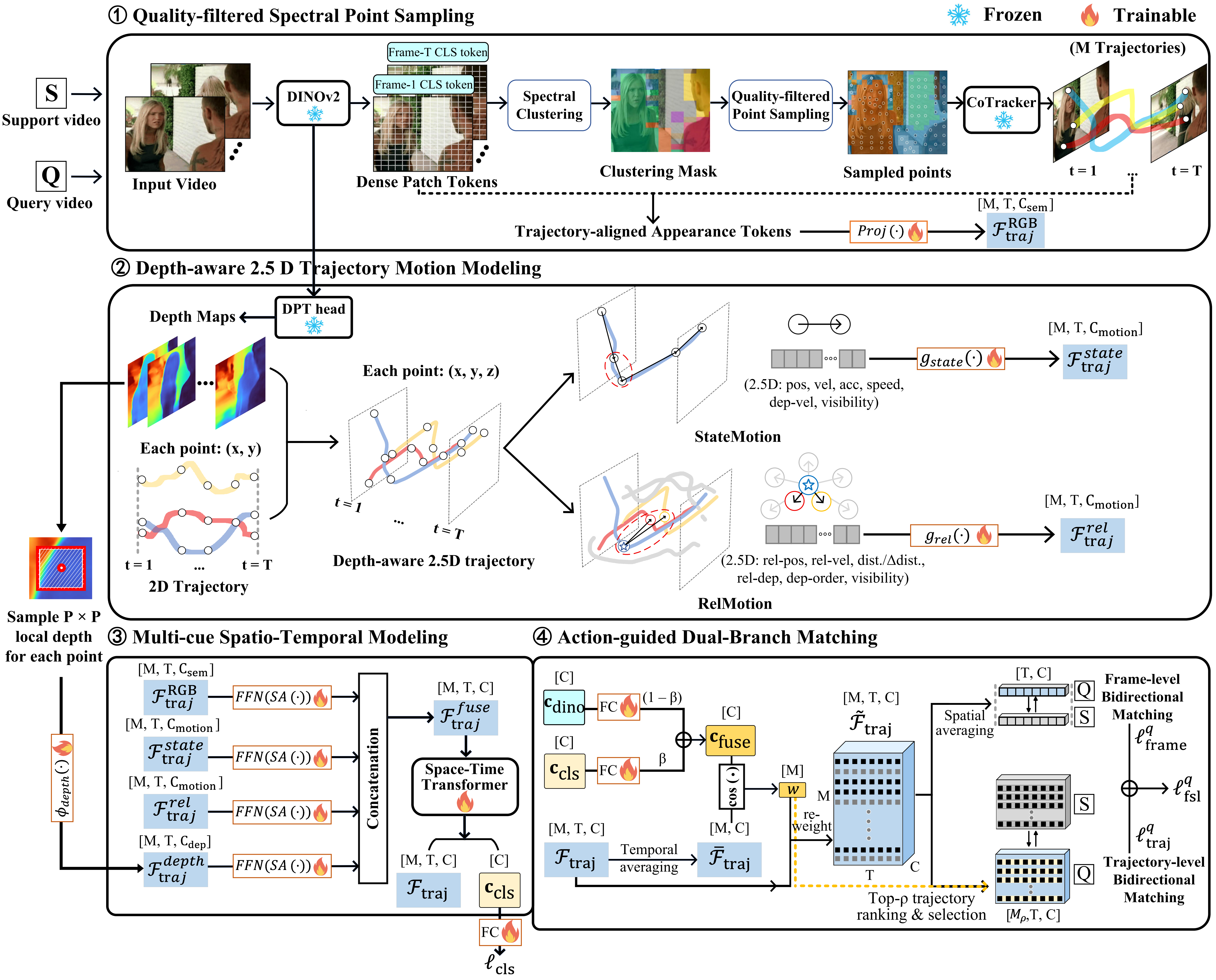}
    \vspace{-2mm}
    \caption{Overview of \textbf{GeoTRAM}. Reliable tracking seeds are obtained through quality-filtered spectral sampling over DINOv2~\cite{oquab2024dinov2} patch features and tracked across frames by CoTracker~\cite{karaev2024cotracker}. Relative monocular depth augments the image-plane tracks into 2.5D trajectories, on which \textit{StateMotion} models the ordered evolution of explicit 2.5D intra-trajectory motion states, while \textit{RelMotion} captures state-conditioned local interactions among neighboring trajectories. Semantic, depth, and motion cues undergo cue-specific temporal refinement before concatenation and joint spatio-temporal modeling. Finally, dual-CLS-guided reweighting suppresses less action-relevant trajectories, followed by the complementary frame-level prototype matching and top-$\rho$ trajectory-level instance matching.}
    \label{fig:framework}
    \vspace{-3mm}
\end{figure*}

\section{Methodology}
\label{sec:methodology}

We propose \textbf{GeoTRAM}, a geometry-aware trajectory state modeling framework for FSAR. Given an input video, GeoTRAM first obtains reliable tracking seeds through quality-filtered spectral sampling over DINOv2 patch features~\cite{oquab2024dinov2}, and CoTracker~\cite{karaev2024cotracker} tracks them across frames. Monocular relative depth then augments the image-plane coordinates to construct 2.5D trajectories. GeoTRAM models their dynamics at two complementary levels: \textit{StateMotion} encodes the ordered evolution of explicit intra-trajectory motion states, while \textit{RelMotion} captures time-varying local interactions through relative geometry and state-conditioned neighbor aggregation. Semantic, depth, and motion cues are temporally refined in dedicated low-dimensional subspaces and concatenated to form multi-cue trajectory tokens, which are then processed by a divided space--time Transformer for joint spatio-temporal modeling. Finally, dual-CLS-guided trajectory relevance estimation supports reweighting and top-\(\rho\) selection, followed by complementary frame-level prototype and trajectory-level instance matching for few-shot classification. Figure~\ref{fig:framework} presents the overall framework.

\vspace{-4mm}
\paragraph{Problem Definition.}
\label{sec:problem_def}
Few-shot action recognition aims to recognize novel action classes from only a few labeled videos. We consider the standard few shot setting with disjoint meta-training and meta-testing action classes. Let $\mathcal{D}_{\mathrm{train}}=\{(v_i,y_i)\mid y_i\in\mathcal{C}_{\mathrm{train}}\}$ and $\mathcal{D}_{\mathrm{test}}=\{(v_i,y_i)\mid y_i\in\mathcal{C}_{\mathrm{test}}\}$  denote the meta-training and meta-testing sets, where $\mathcal{C}_{\mathrm{train}}\cap\mathcal{C}_{\mathrm{test}}=\varnothing$.   Each episode defines an $N$-way $K$-shot task, containing a support set $\mathcal{S}$ with $K$ labeled videos from each of $N$ classes and a query set $\mathcal{Q}$ whose videos are classified among the same $N$ episode classes.
\subsection{Quality-filtered Spectral Point Sampling}
\label{subsec:pointsampling}

We construct reliable tracking seeds through quality-filtered spectral point sampling, which combines spectral clustering, cluster-quality filtering, and budgeted point allocation. Given DINOv2~\cite{oquab2024dinov2} patch features
$\mathcal{F}^{\mathrm{RGB}}\in\mathbb{R}^{T\times H_p\times W_p\times C}$,
we perform per-frame spectral clustering using a kNN affinity graph and its normalized graph Laplacian. The resulting cluster centroids are associated across time to form video-level semantic prototypes and assign temporally consistent patch labels. For each semantic cluster, we select its peak frame, defined as the frame in which the cluster occupies the largest number of patches, and initialize its tracking seeds only in that frame. We suppress temporally unstable, isolated, and fragmented regions, while allocating more seeds to spatially compact clusters.
Further details are provided in Supp.Sec.~\ref{supp:quality-filtered-sampling}.
The resulting tracking seeds are $\mathcal{S}_p=\left\{(t_i^{\mathrm{init}},x_i,y_i)\right\}_{i=1}^{M},$
where $(x_i,y_i)$ is the image-plane coordinate of the $i$-th seed and $t_i^{\mathrm{init}}$ is its initialization frame. Given the input video $\mathcal{X}$, we track these seeds using CoTracker ~\cite{karaev2024cotracker} $\mathcal{T}$:$
\mathcal{P},\mathcal{V}=\mathcal{T}(\mathcal{X},\mathcal{S}_p),$
where
$\mathcal{P}=\{\mathcal{P}^{m}\}_{m=1}^{M}$
denotes the predicted point trajectories and $\mathcal{V}$ denotes their visibility masks. Each trajectory
$\mathcal{P}^{m}=[(x_t^m,y_t^m)]_{t=1}^{T}
\in\mathbb{R}^{T\times2}$
records the image-plane location of the $m$-th point across $T$ frames.

\subsection{Depth-aware 2.5D Trajectory Motion Modeling}
\label{subsec:method_motion}

\paragraph{2.5D Trajectory Construction.}

The raw trajectories contain only 2D image-plane coordinates and therefore omit  depth-axis variation. For each frame, we estimate the monocular depth map $D_t$ using a frozen DPT-based depth-estimation head~\cite{ranftl2021vision} trained on top of intermediate DINOv2 features~\cite{oquab2024dinov2}. We sample the estimated depth at each tracked location:
${z}_t^m=\operatorname{SampleDepth}(D_t,x_t^m,y_t^m)$.
The sampled depths are min--max normalized per video to the same range as the normalized image coordinates, producing the depth-aware 2.5D trajectories:
\vspace{-1mm}
\begin{equation}
\begin{aligned}
\tilde{\mathcal{P}} = [\mathbf{p}_t^m]_{m=1, t=1}^{M, T} \in \mathbb{R}^{M \times T \times 3}, \quad \\
\text{where } \mathbf{p}_t^m = (x_t^m, y_t^m, z_t^m) \in \mathbb{R}^{3}.
\label{eq:3dcoordinate}
\end{aligned}
\vspace{-2mm}
\end{equation}

\vspace{-4mm}
\paragraph{StateMotion: intra-trajectory state modeling.}
We first model the motion state evolution of each individual trajectory. Let $v_t^m\in\{0,1\}$ denote the visibility of trajectory $m$ at time $t$. We compute visibility-aware differences for each lifted trajectory $\tilde{\mathcal{P}}^m$ (Eq. \ref{eq:3dcoordinate}):
\vspace{-1mm}
\begin{equation}
\Delta\mathbf{p}_t^m=
\begin{cases}
\mathbf{p}_t^m-\mathbf{p}_{t-1}^m, & v_t^m v_{t-1}^m=1,\\
\mathbf{0}, & \text{otherwise}.
\end{cases}
\label{eq:vis_differences}
\end{equation}

The acceleration $\Delta^2\mathbf{p}_t^m$ is computed by applying the same mechanism to the velocity.
For each trajectory at each timestep, we construct a 13-dim motion state vector:
\vspace{-2mm}
\begin{equation}
\begin{aligned}
\mathbf{s}_t^m=
[
\mathbf{p}_t^m,\;
\Delta\mathbf{p}_t^m,\;
\Delta^2\mathbf{p}_t^m,\; 
\|\Delta\mathbf{p}_{t,xy}^m\|,\\
\|\Delta\mathbf{p}_t^m\|,\;
\Delta z_t^m,\;
v_t^m
]\in\mathbb{R}^{13}.
\end{aligned}
\label{eq:intra_statevector}
\end{equation}
This vector consists of 2.5D position $(x,y,z)$, velocity $(dx,dy,dz)$, acceleration $(ddx,ddy,ddz)$, 2D/2.5D speed, depth-axis velocity, and visibility, where $\Delta\mathbf{p}_{t,xy}^m$ denotes the image-plane component of $\Delta\mathbf{p}_t^m$ and $\Delta z_t^m$ denotes its depth component. These components provide complementary cues: position encodes the trajectory's instantaneous 2.5D location, velocity and acceleration characterize first- and second-order dynamics, 2D/2.5D speeds encode motion magnitude, $\Delta z_t^m$ explicitly highlights motion along the depth axis, and visibility distinguishes valid observations from occluded or unreliable ones.

We feed each trajectory-level motion state sequence into a lightweight temporal encoder $g_{state}$, which consists of a two-layer temporal 1D convolutional block:
\vspace{-2mm}
\begin{equation}
\mathbf{h}_{1:T}^m
= g_{\text{state}}\left(
[\mathbf{s}_t^m]_{t=1}^{T}
\right)
\in \mathbb{R}^{T \times C_{motion}},
\label{eq:intra_motion_traj}
\end{equation}
where the encoder is shared across all trajectories. The intra-trajectory motion features are then collected as
\vspace{-1mm}
\begin{equation}
\mathcal{F}_{\text{traj}}^{\text{state}}
=
[\mathbf{h}_t^m]_{m=1,t=1}^{M,T}
\in\mathbb{R}^{M\times T \times C_{motion}}.
\label{eq:intra_motion}
\end{equation}
Thus, StateMotion preserves the ordered sequence of explicit motion states and encodes their depth-aware temporal evolution along each trajectory.

\vspace{-5mm}
\paragraph{RelMotion: local motion relation modeling.}
RelMotion further models local interactions among neighboring trajectories. Instead of relating each trajectory to all others, we construct a local trajectory graph that restricts relation modeling to spatially nearby trajectories. For each lifted trajectory $\tilde{\mathcal{P}}^m$ (Eq.\ref{eq:3dcoordinate}), we compute its visibility-weighted temporal 2.5D center by averaging valid trajectory positions over time. We then select its $k$ nearest neighbors $\mathcal{N}_k(m)$ based on Euclidean distance between these centers. The neighborhood graph is constructed once for each video, while relation modeling is performed at every frame.

For each center trajectory $m$ and neighbor $j\in\mathcal{N}_k(m)$, we construct a 12-dim relation state vector at each frame:
\begin{equation}
\begin{aligned}
\mathbf{e}_{t}^{m,j} =
[
&\Delta\mathbf{p}_{t}^{m,j},\;
\Delta\mathbf{u}_{t}^{m,j},\;
d_{xy,t}^{m,j},\;
d_{2.5d,t}^{m,j},\;
\Delta d_{2.5d,t}^{m,j},\\
&\Delta z_t^{m,j},\;
\mathrm{sgn}(\Delta z_t^{m,j}),\;
v_t^m v_t^j
]\in\mathbb{R}^{12}.
\end{aligned}
\label{eq:inter_statevector}
\vspace{-1mm}
\end{equation}

Here, $\Delta\mathbf{p}_{t}^{m,j}=\mathbf{p}_t^j-\mathbf{p}_t^m$ and $\Delta\mathbf{u}_{t}^{m,j}=\mathbf{u}_t^j-\mathbf{u}_t^m$ denote relative 2.5D position and velocity, where $\mathbf{u}_t^m=\Delta\mathbf{p}_t^m$; the terms $d_{xy,t}^{m,j}$ and $d_{2.5d,t}^{m,j}$ measure pairwise 2D and 2.5D distances, while $\Delta d_{2.5d,t}^{m,j}$ captures whether the trajectories approach or separate; $\Delta z_t^{m,j}$ and its sign encode relative depth and depth ordering, and $v_t^m v_t^j$ indicates joint visibility. Together, these cues describe local geometric configuration and relative motion between neighboring trajectories.

Next, since not all neighboring relations are equally informative for action recognition, and their importance should be conditioned on the current motion state of the center trajectory, we encode the relation state vector with an MLP $\phi(\cdot)$ and compute a center-conditioned neighbor attention score using the intra-trajectory motion feature produced by StateMotion:
\vspace{-1mm}
\begin{equation}
\gamma_t^{m,j}
=
a\left(\operatorname{Concat}(\mathbf{h}_t^m,\phi(\mathbf{e}_t^{m,j}))\right),
\label{eq:rel_attn_logits}
\vspace{-1mm}
\end{equation}
where $\mathbf{h}_t^m=
\left[\mathcal{F}_{\text{traj}}^{\text{state}}\right]_{m,t}$
(Eqs.~\ref{eq:intra_motion_traj} and~\ref{eq:intra_motion})
denotes the intra-trajectory motion feature of trajectory $m$ at frame $t$,
and $a(\cdot)$ is a lightweight attention scoring network.

Invalid relations with $v_t^m v_t^j=0$ are masked, and the remaining scores are normalized over $\mathcal{N}_k(m)$ with softmax to obtain the attention weights $\alpha_t^{m,j}$. If no valid neighbor exists at a frame, aggregated relation feature is set to zero.

The relation feature is then obtained by attentive aggregation over valid neighbor relations:
\vspace{-1mm}
\begin{equation}
\mathbf{r}_t^m=
\sum_{j\in\mathcal{N}_k(m)}
\alpha_t^{m,j}\phi(\mathbf{e}_t^{m,j}).
\label{eq:attn_aggregation}
\end{equation}
\vspace{-1mm}
The shared projection head $g_{\text{rel}}(\cdot)$ maps each $\mathbf{r}_t^m$ to the motion dimension, and the projected features are collected as
\begin{equation}
\mathcal{F}_{\text{traj}}^{\text{rel}}
=
[g_{\text{rel}}(\mathbf{r}_t^m)]_{m=1,t=1}^{M,T}
\in\mathbb{R}^{M\times T\times C_{motion}}.
\label{eq:inter_motion}
\end{equation}
Thus, RelMotion captures state-conditioned local interaction dynamics by explicitly modeling the relative motion, proximity changes, and depth relations of neighboring trajectories in the 2.5D trajectory space.

\subsection{Multi-cue Spatio-Temporal Modeling}
\label{sec:multicue_fusion}

For each tracked point $m$ at frame $t$, we construct a multi-cue trajectory token from semantic, depth, intra-motion, and inter-motion features. The RGB semantic feature map $\mathcal{F}^{\text{RGB}}$ is aligned with the 2D trajectories ${\mathcal{P}}$ via nearest-neighbor sampling~\cite{kumar2024tats} and projected to $C_{sem}$ dimensions:
\begin{equation}
\mathcal{F}_{\text{traj}}^{\text{RGB}}
=
\operatorname{Proj}(\operatorname{Align}(\mathcal{F}^{\text{RGB}},{\mathcal{P}}))
\in\mathbb{R}^{M\times T\times C_{sem}}.
\end{equation}

Then, the depth map $D_{t}$ in frame $t$ is sampled bilinearly with the local $P \times P$  sampling grid $\mathcal{G}_{P}$ centered on $(x_t^m,y_t^m)$, producing $\mathbf{D}_{t,m}^{\text{patch}}$. It is further mapped into $C_{dep}$-dim depth feature using a lightweight depth encoder:

\begin{equation}
\mathcal{F}_{\text{traj}}^{\text{depth}}
=
\left[
\phi_{\text{depth}}\big(\mathbf{D}_{t,m}^{\text{patch}}\big)
\right]_{m=1,t=1}^{M,T}
\in\mathbb{R}^{M\times T\times C_{\text{dep}}}.
\end{equation}
In total,
we obtain four cue-specific representations:
$\mathcal{F}_{\mathrm{traj}}^{\mathrm{RGB}}$,
$\mathcal{F}_{\mathrm{traj}}^{\mathrm{state}}$,
$\mathcal{F}_{\mathrm{traj}}^{\mathrm{rel}}$, and
$\mathcal{F}_{\mathrm{traj}}^{\mathrm{depth}}$.
Each cue is processed by temporal self-attention and a
feed-forward network, and the resulting features are concatenated along
the channel dimension to form
$\mathcal{F}_{\mathrm{traj}}^{\mathrm{fuse}}\in\mathbb{R}^{M\times T\times C}$, where $C=C_{\mathrm{sem}}+C_{\mathrm{dep}}+2C_{\mathrm{motion}}$,
preserving cue-specific information before joint spatio-temporal modeling.

The resulting multi-cue trajectory tokens $\mathcal{F}_{\mathrm{traj}}^{\mathrm{fuse}}$ are then processed by a one-block divided space--time Transformer with a learnable class token, following~\cite{kumar2024tats}. The visibility masks from CoTracker exclude invalid trajectory observations from attention. The Transformer outputs a global representation $\mathbf{c}_{\mathrm{cls}}\in\mathbb{R}^{C}$ and trajectory embeddings
\begin{equation}
\mathcal{F}_{\mathrm{traj}}
=[\mathbf{f}_t^m]_{m=1,t=1}^{M,T}
\in\mathbb{R}^{M\times T\times C},
\label{eq:final_tokens}
\end{equation}
which are subsequently used for few-shot classification.

\subsection{Action-guided Dual-Branch Matching}
\label{subsec:classification}

\paragraph{Dual-CLS-guided trajectory relevance.} Beyond its use for auxiliary global supervision, the video-level class token is used as a global action representation to estimate trajectory relevance. The video-level class token summarizes global action context from the fused trajectory representation, while the DINOv2
CLS token representation provides complementary semantic appearance context. Since sparse point sampling may still retain background or less informative tracks, we fuse these two global representations to construct an action-semantic anchor for trajectory reweighting. Specifically,
\(\mathbf c_{\mathrm{dino}}\) is obtained by temporally averaging the frame-wise DINOv2 CLS tokens. Together with the video-level class token $\mathbf{c}_{\mathrm{cls}}\in\mathbb{R}^{C}$ from
Sec.~\ref{sec:multicue_fusion}, we compute
\vspace{-1mm}
\begin{equation}
\mathbf{c}_{\text{fuse}}
=
\beta\,\phi_v(\mathbf{c}_{\text{cls}})
+
(1-\beta)\,\phi_d(\mathbf{c}_{\text{dino}}),
\label{eq:fused_cls}
\end{equation}
where $\phi_v$ and $\phi_d$ are learnable projections into a shared C-dimensional space, and
$\beta$ controls their relative contribution. The relevance of
trajectory $m$ is measured by
\vspace{-1mm}
\begin{equation}
s_m=
\cos\!\left(
\mathbf{c}_{\mathrm{fuse}},
\bar{\mathbf{f}}_m
\right),
\label{eq:traj_sim}
\end{equation}
where $\bar{\mathbf{f}}_m$ is the visibility-weighted temporal average of the
tokens $[\mathbf{f}_t^m]_{t=1}^{T}$ for trajectory $m$, producing $\bar{\mathcal{F}}_{\mathrm{traj}}=[\bar{\mathbf f}_m]_{m=1}^M\in\mathbb{R}^{M\times C}$. We convert $s_m$ into
a soft relevance weight $w_m=\sigma(s_m/\tau)$, where $\tau$ is a temperature
parameter, and reweight each trajectory token as
$\tilde{\mathbf{f}}_t^m=w_m\mathbf{f}_t^m$.
The resulting representations form
$\tilde{\mathcal{F}}_{\mathrm{traj}}
\in\mathbb{R}^{M\times T\times C}$
for subsequent dual-branch few-shot classification.

\vspace{-3mm}
\paragraph{Frame-level prototype classification.}
Given $\tilde{\mathcal{F}}_{\mathrm{traj}}$, we adopt a dual-branch
few-shot classifier. The frame-level branch first averages the reweighted
trajectory tokens over trajectories to obtain $\mathbf{g}_t
=
\frac{1}{M}
\sum_{m=1}^{M}
\tilde{\mathbf{f}}_t^m.$
For each class, the frame representations of its $K$ support videos are
further averaged to form a class prototype sequence. A query is then matched
to each class prototype using the bidirectional set-matching metric introduced
in prior works~\cite{wang2022hyrsm,wang2024hyrsm++} and widely adopted in subsequent FSAR
methods~\cite{wang2022hyrsm, wang2024hyrsm++, wang2023molo,kumar2024tats,kumar2025trokens}, producing the
frame-level logits $\boldsymbol{\ell}_{\mathrm{frame}}^{q}$.

\vspace{-3.5mm}
\paragraph{Trajectory-native instance classification.}
The complementary trajectory branch preserves trajectory-level structure.
For each video, we retain the top-$\rho$ fraction of trajectories ranked by
$w_m$, selecting $M_\rho=\lfloor\rho M\rfloor$  trajectories. The selected trajectories are retained across all $T$ frames. For a selected query trajectory $m_q$ and support trajectory $m_s$, we compute a time-averaged trajectory similarity:
\vspace{-2mm}
\begin{equation}
\bar{s}^{q,s}(m_q,m_s)
=
\frac{1}{T}\sum_{t=1}^{T}
\cos\bigl(
\tilde{\mathbf{f}}_{t}^{q,m_q},
\tilde{\mathbf{f}}_{t}^{s,m_s}
\bigr).
\label{eq:traj_pair_sim}
\vspace{-1mm}
\end{equation}
Bidirectional max--mean matching over the two trajectory sets then gives
the instance score
$\mathrm{score}_{q,s}
=
\tfrac12(s_{q\rightarrow s}+s_{s\rightarrow q})$,
where
$s_{q\rightarrow s}
=
\frac{1}{M_\rho}
\sum_{m_q}\max_{m_s}\bar{s}^{q,s}(m_q,m_s)$,
and $s_{s\rightarrow q}$ is defined symmetrically, with the sums and maxima
taken over the selected trajectories.
Class logits $\boldsymbol{\ell}_{\mathrm{traj}}^{q}$ are obtained by class-wise score averaging.
Finally, the two complementary branches are combined as
\begin{equation}
\boldsymbol{\ell}_{\mathrm{fsl}}^{q}
=
\boldsymbol{\ell}_{\mathrm{frame}}^{q}
+
\boldsymbol{\ell}_{\mathrm{traj}}^{q}.
\end{equation}

\vspace{-3mm}
\paragraph{Training objective.}
Following prior FSAR methods~\cite{wang2023molo,wang2022hyrsm,
wang2024hyrsm++,kumar2024tats,kumar2025trokens}, we apply an auxiliary
meta-class classifier to the Transformer class token,
$\boldsymbol{\ell}_{\mathrm{cls}}
=\operatorname{FC}(\mathbf{c}_{\mathrm{cls}})
\in\mathbb{R}^{|C_{\mathrm{train}}|}$.
The model is trained using the sum of the auxiliary classification loss
and the episodic classification loss:
\vspace{-0.5mm}
\begin{equation}
\mathcal{L}
=
\mathcal{L}_{\mathrm{CE}}
(\boldsymbol{\ell}_{\mathrm{cls}}^{Q},y^{\mathrm{cls}})
+
\mathcal{L}_{\mathrm{CE}}
(\boldsymbol{\ell}_{\mathrm{fsl}}^{Q},y^{\mathrm{epi}}).
\label{eq:loss}
\end{equation}

\section{Experiments}
\label{sec:results}

\begin{table*}[!t]
\centering
\renewcommand{\arraystretch}{0.80}
\caption{Comparison of few-shot action accuracy (\%) under 1--5 shot settings on SSv2-Full and Kinetics. Best results are in \textbf{bold}, second-best results are \underline{underlined}.  A more comprehensive version of the table is shown in the supplementary material (Supp.Sec.\ref{sec:supp-tables}).}
\vspace{-1mm}
\resizebox{0.9\textwidth}{!}{
\begin{tabular}{lcccccccccccc}
\toprule
\textbf{Method} & \textbf{Venue} &
\multicolumn{5}{c}{\textbf{SSv2-Full}} &
\multicolumn{5}{c}{\textbf{Kinetics}} \\
\cmidrule(lr){3-7}\cmidrule(lr){8-12}
& & 1-shot & 2-shot & 3-shot & 4-shot & 5-shot
  & 1-shot & 2-shot & 3-shot & 4-shot & 5-shot \\
\midrule
OTAM~\cite{cao2020otam} & CVPR'20 & 42.8 & 49.1 & 51.5 & 52.0 & 52.3 & 72.2 & 75.9 & 78.7 & 81.9 & 84.2 \\
TRX~\cite{perrett2021trx} & CVPR'21 & 42.0 & 53.1 & 57.6 & 61.1 & 64.6 & 63.6 & 76.2 & 81.8 & 83.4 & 85.2 \\
STRM~\cite{thatipelli2022strm} & CVPR'22 & 43.1 & 53.3 & 59.1 & 61.7 & 68.1 & 62.9 & 76.4 & 81.1 & 83.8 & 86.7 \\
HyRSM~\cite{wang2022hyrsm} & CVPR'22 & 54.3 & 62.2 & 65.1 & 67.9 & 69.0 & 73.7 & 80.0 & 83.5 & 84.6 & 86.1 \\
MoLo~\cite{wang2023molo} & CVPR'23 & 56.6 & 62.3 & 67.0 & 68.5 & 70.6 & 74.0 & 80.4 & 83.7 & 84.7 & 85.6 \\
RFPL~\cite{xia2023rfpl} & ICCV'23 & 47.0 & 54.6 & 58.3 & 60.3 & 61.0 & 74.6 & 80.0 & 82.1 & 84.1 & 86.8 \\
HyRSM++~\cite{wang2024hyrsm++} & PR'24 & 55.0 & 63.5 & 66.0 & 68.8 & 69.8 & 74.0 & 80.8 & 83.9 & 85.3 & 86.4 \\
TATs~\cite{kumar2024tats} & ECCV'24 & 57.7 & 67.1 & 70.0 & 70.6 & 74.6 & 81.9 & 86.5 & \underline{89.9} & 90.6 & 91.1 \\
TEAM~\cite{lee2025team} & CVPR'25 & -- & -- & -- & -- & -- & \textbf{83.3} & -- & -- & -- & \textbf{92.9} \\
Trokens~\cite{kumar2025trokens} & ICCV'25 & \underline{61.5} & \underline{69.9} & \underline{73.8} & \underline{75.9} & \underline{76.7} & 82.9 & \textbf{87.7} & \underline{89.9} & \underline{90.8} & 91.2 \\

\midrule
\textbf{GeoTRAM} & - & \textbf{68.9} & \textbf{76.5} & \textbf{79.9} & \textbf{81.9} & \textbf{82.3} & \underline{83.1} & \underline{87.2} & \textbf{90.2} & \textbf{91.0} & \underline{91.8} \\

\bottomrule
\end{tabular}}
\label{tab:table1}
\vspace{-1mm}
\end{table*}


\begin{table*}[!t]
\centering
\renewcommand{\arraystretch}{0.80}
\scriptsize
\caption{Comparison of few-shot action accuracy (\%) under 1-, 3-, and 5-shot settings on SSv2-Small, UCF-101, and HMDB-51. Best results are in \textbf{bold}, second-best results are \underline{underlined}.  A more comprehensive version of the table is shown in the Supp.Sec.\ref{sec:supp-tables}.}
\vspace{-1mm}
\resizebox{0.9\textwidth}{!}{
\begin{tabular}{lcccccccccc}
\toprule
\textbf{Method} & \textbf{Venue} &
\multicolumn{3}{c}{\textbf{SSv2-Small}} &
\multicolumn{3}{c}{\textbf{UCF-101}} &
\multicolumn{3}{c}{\textbf{HMDB-51}} \\
\cmidrule(lr){3-5}\cmidrule(lr){6-8}\cmidrule(lr){9-11}
& & 1-shot & 3-shot & 5-shot
  & 1-shot & 3-shot & 5-shot
  & 1-shot & 3-shot & 5-shot \\
\midrule
OTAM~\cite{cao2020otam} & CVPR'20 & 36.4 & 45.9 & 48.0 & 79.9 & 87.0 & 88.9 & 54.5 & 65.7 & 68.0 \\
TRX~\cite{perrett2021trx} & CVPR'21 & 36.0 & 51.9 & 56.7 & 78.2 & 92.4 & 96.1 & 53.1 & 66.8 & 75.6 \\
STRM~\cite{thatipelli2022strm} & CVPR'22 & 37.1 & 49.2 & 55.3 & 80.5 & 92.7 & 96.9 & 52.3 & 67.4 & 77.3 \\
HyRSM~\cite{wang2022hyrsm} & CVPR'22 & 40.6 & 52.3 & 56.1 & 83.9 & 93.0 & 94.7 & 60.3 & 71.7 & 76.0 \\
MoLo~\cite{wang2023molo} & CVPR'23 & 42.7 & 52.9 & 56.4 & 86.0 & 93.5 & 95.5 & 60.8 & 72.0 & 77.4 \\
HyRSM++~\cite{wang2024hyrsm++} & PR'24 & 42.8 & 52.4 & 58.0 & 85.8 & 93.5 & 95.9 & 61.5 & 72.7 & 76.4 \\
TATs~\cite{kumar2024tats} & ECCV'24 & 47.9 & 60.0 & 64.4 & 92.0 & 96.8 & 95.5 & 60.0 & 71.8 & 77.0 \\
TEAM~\cite{lee2025team} & CVPR'25 & 47.2 & -- & 63.1 & \textbf{94.5} & -- & \textbf{98.8} & \underline{70.9} & -- & \textbf{85.5} \\
Trokens~\cite{kumar2025trokens} & ICCV'25 & \underline{53.4} & \underline{65.3} & \underline{68.9} & \underline{94.0} & \underline{97.3} & 97.9 & 69.8 & \underline{80.0} & 82.3 \\

\midrule
\textbf{GeoTRAM} & - & \textbf{57.7} & \textbf{70.4} & \textbf{74.1} & 93.5  & \textbf{97.5} & \underline{98.2} & \textbf{71.2} & \textbf{81.0} & \underline{83.9} \\
\bottomrule
\end{tabular}}
\label{tab:table2}
\vspace{-1mm}
\end{table*}


\begin{table*}[!t]
\renewcommand{\arraystretch}{0.80}
\scriptsize
\caption{Comparison of N-way 1-shot classification accuracy (N=5--10) on SSv2-Full and Kinetics datasets with recent representative FSAR methods. Best and second-best results are in \textbf{bold} and \underline{underlined}, respectively.}
\vspace{-1mm}
\resizebox{\textwidth}{!}{
\begin{tabular}{@{}lcccccc|cccccc@{}}

\toprule
& \multicolumn{6}{c}{\textbf{SSv2-Full}} & \multicolumn{6}{c}{\textbf{Kinetics}}  \\ 
\cmidrule(l){2-7} \cmidrule(l){8-13} 

\textbf{Method} &  \multicolumn{1}{l}{5-way} & \multicolumn{1}{l}{6-way} & \multicolumn{1}{l}{7-way} & \multicolumn{1}{l}{8-way} & \multicolumn{1}{l}{9-way} & \multicolumn{1}{l}{10-way}  &  \multicolumn{1}{l}{5-way} & \multicolumn{1}{l}{6-way} & \multicolumn{1}{l}{7-way} & \multicolumn{1}{l}{8-way} & \multicolumn{1}{l}{9-way} & \multicolumn{1}{l}{10-way} \\
\midrule
OTAM~\cite{cao2020otam} & 42.8 & 38.6 & 35.1 & 32.3 & 30.0 & 28.2 & 72.2 & 68.7 & 66.0 & 63.0 & 61.9 & 59.0 \\
TRX~\cite{perrett2021trx} & 42.0 & 41.5 & 36.1 & 33.6 & 32.0 & 30.3 & 63.6 & 59.4 & 56.7 & 54.6 & 53.2 & 51.1 \\
HyRSM~\cite{wang2022hyrsm} & 54.3 & 50.1 & 45.8 & 44.3 & 42.1 & 40.0 & 73.7 & 69.5 & 66.6 & 65.5 & 63.4 & 61.0 \\
MoLo~\cite{wang2023molo} & 56.6 & 51.6 & 48.1 & 44.8 & 42.5 & 40.3 & 74.0 & 69.7 & 67.4 & 65.8 & 63.5 & 61.3 \\
TATs~\cite{kumar2024tats} & 57.7 & 55.7 & 52.5 & 50.0 & 47.0 & 45.8 & 81.9 & 79.0 & 76.1 & 75.2 & 72.2 & \underline{72.0} \\
Trokens~\cite{kumar2025trokens} & \underline{61.5} & \underline{59.1} & \underline{56.5} & \underline{54.6} & \underline{51.4} & \underline{49.1} & \underline{82.9} & \underline{80.2} & \textbf{78.5} & \textbf{76.8} & \textbf{75.5} & \textbf{73.3} \\
\midrule
\textbf{GeoTRAM} & \textbf{68.9} & \textbf{65.0} & \textbf{61.3} & \textbf{59.4} & \textbf{55.8} & \textbf{54.1} & \textbf{83.1} & \textbf{80.5} & \underline{77.5} & \underline{75.9} & \underline{74.3} & 71.8 \\
\bottomrule
\end{tabular}}
\label{tab:table3}
\vspace{-2mm}
\end{table*}

\subsection{Datasets and few-shot setup}

We evaluate our approach on standard FSAR benchmarks with few-shot splits: SSv2 (Something-Something)~\cite{goyal2017something}, Kinetics~\cite{carreira2017quo}, UCF-101~\cite{ucf101}, and HMDB-51~\cite{hmdb51}. For SSv2, we follow the standard 64/12/24 class split for meta-training/validation/testing~\cite{zhu2018compound,cao2020otam}; SSv2-Small contains 100 videos per class, whereas SSv2-Full uses all available videos per class. We use the Kinetics splits from~\cite{huang2022compound} and the UCF-101 and HMDB-51 splits from~\cite{wu2022motion,zhang2020few}, following prior FSAR protocols~\cite{wang2023molo,kumar2024tats,kumar2025trokens}.

\subsection{Implementation details}
The DINOv2 ViT-B/14 backbone~\cite{oquab2024dinov2}, its DPT-based~\cite{ranftl2021vision} pretrained depth-estimation head, and CoTracker~\cite{karaev2024cotracker} are kept frozen, while the single-block space--time Transformer and remaining modules are optimized using Eq.~\ref{eq:loss}. Point trajectories are precomputed offline with CoTracker, following baseline methods~\cite{kumar2024tats, kumar2025trokens}, whereas depth maps are estimated online during training and evaluation. Following recent FSAR protocols~\cite{wang2023molo,kumar2024tats,kumar2025trokens}, we uniformly sample $T=8$ frames and track $M=256$ points per video, train for 50 epochs using episodic meta-learning, and evaluate over 10{,}000 test episodes. We set the RelMotion neighborhood size $k=8$, local depth patch size $P=14$, CLS fusion coefficient $\beta=0.5$, and trajectory retention ratio $\rho=0.5$. To match the baseline works~\cite{kumar2024tats, kumar2025trokens} and the ViT-B/14 embedding dimension, we set $C=768$, with $C_{\mathrm{sem}}=384$, $C_{\mathrm{motion}}=176$, and $C_{\mathrm{dep}}=32$. We also introduce GeoTRAM-S, a compact variant designed for lower computational budget, with $C=560$. All experiments are conducted on a single NVIDIA A100-40GB GPU. More implementation details are provided in Supp.Sec.~\ref{sec:supp-impl}.

\subsection{Comparison with State-of-the-Art Methods}

As shown in Tabs.~\ref{tab:table1} and~\ref{tab:table2}, GeoTRAM achieves its largest gains over the closest competitor, Trokens~\cite{kumar2025trokens}, on the motion-centric Something-Something benchmarks. It improves SSv2-Full by 5.6--7.4\% across the 1/3/5-shot settings and SSv2-Small by 4.3/5.1/5.2\%. Beyond Trokens, GeoTRAM also surpasses all other compared methods across every reported setting on both benchmarks, by 7.7--11.3\% on SSv2-Full and 9.7--10.4\% on SSv2-Small over the strongest non-Trokens results. Since Something-Something actions are largely distinguished by temporally ordered motion and hand--object interactions, these gains support the effectiveness of our geometry-aware trajectory state modeling framework for motion-sensitive few-shot recognition. Class-wise analysis in Supp.Sec.~\ref{supp:classwise_analysis} further shows that many of the largest gains occur for actions involving structured object motion and relative spatial relations.

GeoTRAM remains competitive on more appearance-driven benchmarks. It outperforms competitor Trokens in four of the five Kinetics settings, under the 3-/5-shot settings on UCF-101, and across all evaluated settings on HMDB-51. Under the challenging $N$-way protocol in Tab.~\ref{tab:table3}, GeoTRAM consistently improves over Trokens on SSv2-Full from 5-way to 10-way, with gains of 4.4--7.4\%. On Kinetics, it outperforms Trokens at 5- and 6-way but falls below it from 7-way onward, indicating that these gains do not persist as the number of classes increases. Overall, GeoTRAM is particularly effective for motion-centric and interaction-heavy action recognition while maintaining competitive performance on appearance-driven datasets.

\subsection{Ablation studies}
Note that we regard TATs~\cite{kumar2024tats} as a clean ablation baseline and Trokens~\cite{kumar2025trokens} as the closest competitor for direct comparison. GFLOPs exclude the frozen backbone and offline point initialization and tracking, consistent with these two works, while including our online depth estimation. More experimental analyses are provided in the Supp.Sec.\ref{supp:more_ablations}.

\vspace{-5mm}
\paragraph{Component ablations.}
Tab.~\ref{tab:components} evaluates the GeoTRAM components on SSv2-Small and SSv2-Full, starting from the reproduced TATs with uniform sampling and appearance-only spatio-temporal modeling. Semantic-aware sampling improves several settings over uniform sampling but shows inconsistent gains, whereas our quality-filtered spectral sampling consistently improves all four settings by 2.61/1.97\% on SSv2-Small and 1.64/0.93\% on SSv2-Full, supporting more reliable trajectory initialization. StateMotion further provides gains of 2.84/2.34\% and 2.93/1.87 \%, while RelMotion adds another 0.84/0.93\% and 1.51/0.92\%, respectively, supporting the complementary roles of ordered intra-trajectory state evolution and local inter-trajectory interactions. Multi-cue spatio-temporal modeling (STM) produces additional improvements of 1.95/2.61\% and 2.45/2.08\%, demonstrating the benefit of preserving heterogeneous cues for joint spatio-temporal modeling. Finally, Dual-CLS-guided trajectory reweighting and dual-branch matching provide consistent further gains across both datasets. Overall, these results demonstrate the complementary contributions of the proposed components. 

\vspace{-5mm}
\paragraph{Motion--fusion interaction.}
Table~\ref{tab:motion_fusion_interaction} examines the interaction between GeoTRAM (StateMotion+RelMotion) motion modeling and feature fusion on SSv2-Small, with all other components fixed. Under addition, GeoTRAM-motion improves over Trokens-motion by 0.84/1.15/1.28\% in the 1/3/5-shot settings, while the gains increase to 1.40/1.66/1.49\% under multi-cue fusion. Conversely, multi-cue fusion consistently provides larger improvements with GeoTRAM-motion than with Trokens-motion. These results support the complementarity between trajectory-state motion modeling and cue-preserving fusion, with their combination performing best at low computational cost.

\begin{table}[t]
\vspace{-1mm}
\centering
\scriptsize
\setlength{\tabcolsep}{2.2pt}
\renewcommand{\arraystretch}{1.08}
\caption{Component ablation of GeoTRAM on SSv2-Small and SSv2-Full. Starting from the reproduced TATs baseline (first row), we compare semantic-aware and spectral sampling before progressively adding the remaining GeoTRAM components.}
\resizebox{\linewidth}{!}{%
\begin{tabular}{@{}c|cccccccccc@{}}
\hline
\multirow{2}{*}{\makecell{\textbf{Semantic}\\\textbf{Sampling}}} &
\multirow{2}{*}{\makecell{\textbf{Spectral}\\\textbf{Sampling}}} &
\multicolumn{2}{c}{\textbf{Motion Modeling}} &
\multirow{2}{*}{\makecell{\textbf{MultiCue}\\\textbf{STM}}} &
\multirow{2}{*}{\makecell{\textbf{Dual-CLS}\\\textbf{Reweight}}} &
\multirow{2}{*}{\makecell{\textbf{Dual-branch}\\\textbf{Matching}}} &
\multicolumn{2}{c}{\textbf{SSv2-Small}} &
\multicolumn{2}{c}{\textbf{SSv2-Full}} \\
\cline{3-4}
\cline{8-9}
\cline{10-11}
& & \textbf{State.} & \textbf{Rel.} & & & &
1-shot & 5-shot & 1-shot & 5-shot \\
\hline

\cross & \cross & \cross & \cross & \cross & \cross & \cross
& 47.91 & 64.42 & 57.78 & 74.65 \\

\tick & \cross & \cross & \cross & \cross & \cross & \cross
& 49.93 & 65.71 & 58.62 & 74.26 \\

\cross & \tick & \cross & \cross & \cross & \cross & \cross
& 50.52 & 66.39 & 59.42 & 75.58 \\

\cross & \tick & \tick & \cross & \cross & \cross & \cross
& 53.36 & 68.73 & 62.35 & 77.45 \\

\cross & \tick & \tick & \tick & \cross & \cross & \cross
& 54.20 & 69.66 & 63.86 & 78.37 \\

\cross & \tick & \tick & \tick & \tick & \cross & \cross
& 56.15 & 72.27 & 66.31 & 80.45 \\

\cross & \tick & \tick & \tick & \tick & \tick & \cross
& 56.86 & 73.14 & 67.65 & 81.40 \\

\cross & \tick & \tick & \tick & \tick & \tick & \tick
& \textbf{57.71} & \textbf{74.07}
& \textbf{68.85} & \textbf{82.30} \\

\hline
\end{tabular}%
}
\vspace{-1mm}
\label{tab:components}
\end{table}

\begin{table}
\centering
\scriptsize
\setlength{\tabcolsep}{6.0pt}
\renewcommand{\arraystretch}{1.08}
\caption{Motion--fusion interaction on SSv2-Small. Only motion modeling and fusion strategy vary within the same GeoTRAM configuration. GeoTRAM-motion (State.+Rel.)
produces modest gains under addition but larger gains with multi-cue fusion.}
\label{tab:motion_fusion_interaction}
\resizebox{0.48\textwidth}{!}{%
\begin{tabular}{@{}lccccc@{}}
\toprule
\textbf{Motion Modeling} & \textbf{Fusion} & \textbf{GFLOPs} & \textbf{1-shot} & \textbf{3-shot} & \textbf{5-shot} \\
\midrule
Trokens-motion & Addition & 38 & 54.58 & 66.67 & 70.22 \\
GeoTRAM-motion & Addition & 26 & 55.42 & 67.82 & 71.50 \\
\hdashline
Trokens-motion & Multi-cue fusion & 42 & 56.31 & 68.70 & 72.58  \\
GeoTRAM-motion & Multi-cue fusion  & 28 & 57.71 & 70.36 & 74.07 \\
\bottomrule
\end{tabular}%
}
\vspace{-2mm}
\end{table}

\vspace{-4mm}
\paragraph{Efficiency--accuracy trade-off.}
Table~\ref{tab:efficiency} demonstrates a favorable accuracy--efficiency trade-off among trajectory-based methods. Compared with Trokens, GeoTRAM reduces trainable parameters from 17M to 13M and computation from 35 to 28 GFLOPs, while improving accuracy across all shot settings on both datasets. The compact GeoTRAM-S requires only 8M parameters and 17 GFLOPs, outperforms TATs and Trokens across all settings, and even remains close to the full model.

\begin{table}[t]
\centering
\scriptsize
\setlength{\tabcolsep}{3.0pt}
\renewcommand{\arraystretch}{1.08}
\caption{Efficiency--accuracy comparison. We report 5-way accuracy (\%). Param. refers to trainable parameters. GeoTRAM achieves a favorable accuracy--efficiency trade-off over prior trajectory-based FSAR methods.}
\vspace{-1mm}
\resizebox{\columnwidth}{!}{
\begin{tabular}{lcc ccc ccc}
\toprule
\multirow{2}{*}{\textbf{Method}} 
& \multirow{2}{*}{\textbf{Param.}} 
& \multirow{2}{*}{\textbf{GFLOPs}} 
& \multicolumn{3}{c}{\textbf{SSv2-Small}} 
& \multicolumn{3}{c}{\textbf{SSv2-Full}} \\
\cmidrule(lr){4-6} \cmidrule(lr){7-9}
& & & 1-shot & 3-shot & 5-shot & 1-shot & 3-shot & 5-shot \\
\midrule
TATs~\cite{kumar2024tats}
& 11M & 21 
& 47.90 & 60.00 & 64.40 
& 57.70 & 70.00 & 74.60 \\

Trokens~\cite{kumar2025trokens}
& 17M & 35 
& 53.40 & 65.30 & 68.90 
& 61.50 & 73.80 & 76.70 \\
\hdashline
GeoTRAM
& 13M & 28 
& \textbf{57.71} & \textbf{70.36} & \textbf{74.07} 
& \textbf{68.85} & \textbf{79.89} & 82.30 \\

GeoTRAM-S
& 8M & 17 
& 57.04 & 68.40 & 72.78 
& 68.75 & 79.26 & \textbf{83.01} \\
\bottomrule
\end{tabular}
}
\label{tab:efficiency}
\vspace{-5mm}
\end{table}

\begin{figure}[t]
    \centering
    \includegraphics[width=0.99\linewidth]{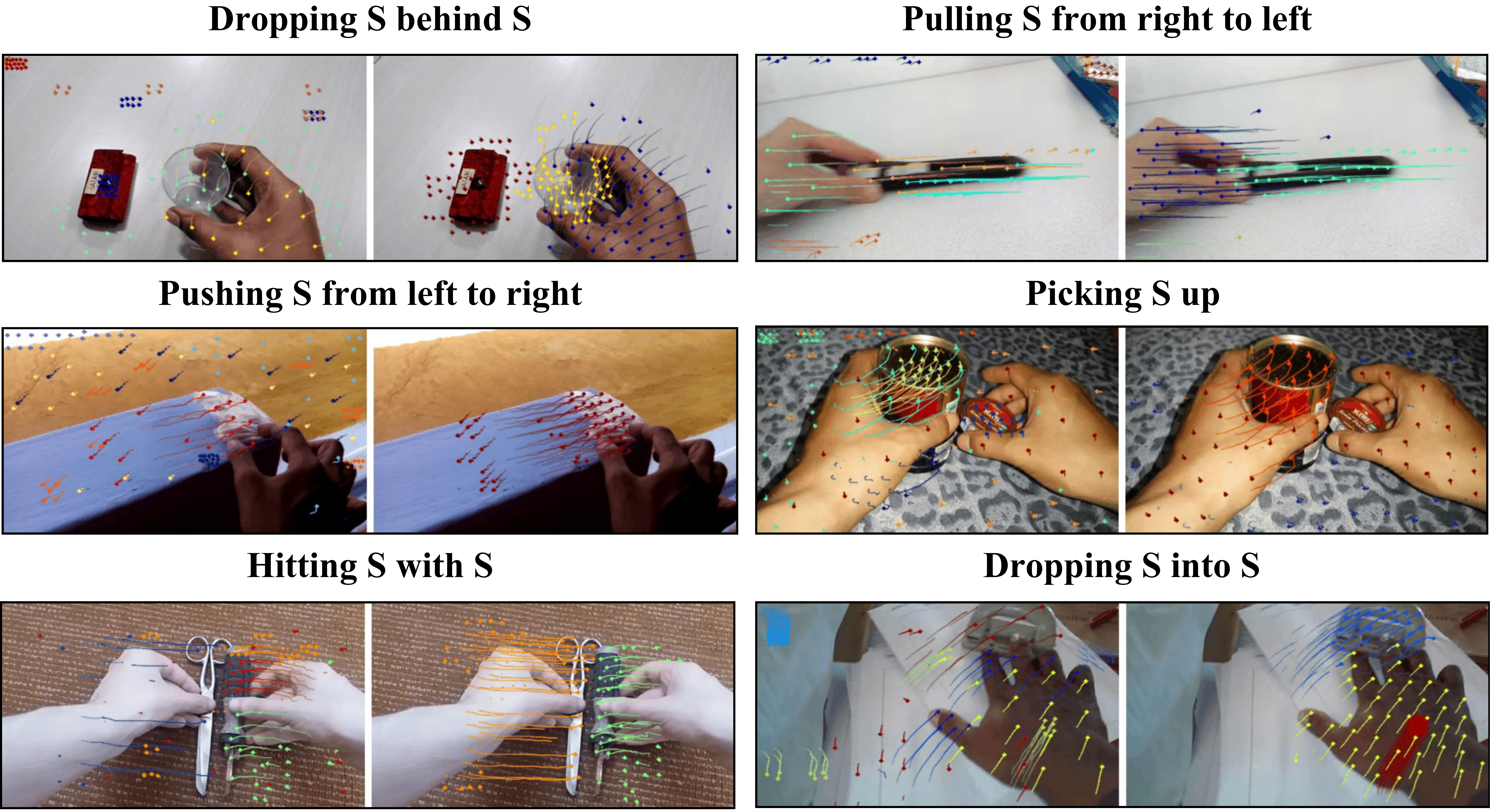}
    \caption{Qualitative comparison of Trokens' (left) and our (right) sampling. Ours reduces scattered background trajectories and concentrates on more spatially coherent action regions.}
    \vspace{-3mm}
    \label{fig:vis_traj_sem_spectral}
\end{figure}

\vspace{-5mm}
\paragraph{Qualitative analysis of point sampling.}
Figure~\ref{fig:vis_traj_sem_spectral} compares our sampling strategy with the semantic-aware baseline. Filtering temporally unstable and spatially fragmented regions while allocating more seeds to compact clusters produces fewer scattered background trajectories and more spatially coherent trajectory groups. Additional comparisons are provided in Supp.Sec.~\ref{supp:point_sampling_comparison}.

\begin{table}[t]
\centering
\scriptsize
\setlength{\tabcolsep}{3.0pt}
\renewcommand{\arraystretch}{1.08}
\caption{Contribution of relative depth to GeoTRAM. ``2D'' removes all depth-related parts, while ``2.5D'' denotes the full model.}
\vspace{-1mm}
\label{tab:depth_ablation}
\begin{tabular}{lcccc}
\toprule
\multirow{2}{*}{\textbf{Representation}}
& \multirow{2}{*}{\textbf{GFLOPs}}
& \multicolumn{3}{c}{\textbf{SSv2-Small}} \\
\cmidrule(lr){3-5}
& & 1-shot & 3-shot & 5-shot \\
\midrule
GeoTRAM-2D
& 27
& 56.86 & 69.12 & 72.75  \\

GeoTRAM-2.5D
& 28
& 57.71 & 70.36 & 74.07 \\
\bottomrule
\end{tabular}
\vspace{-4mm}
\end{table}

\vspace{-5mm}
\paragraph{Effect of relative depth.}
Table~\ref{tab:depth_ablation} compares GeoTRAM with a 2D variant that removes all depth-related cues. Introducing relative depth improves 1/3/5-shot accuracy by 0.85/1.24/1.32\% with only 1 additional GFLOP. These gains show that depth-axis variation provides complementary geometric information beyond image-plane motion. Notably, even the 2D variant still outperforms Trokens by 3.46/3.82/3.85\%, indicating that GeoTRAM's improvements do not rely solely on the additional depth cue.

\section{Conclusion}
\label{sec:conclusion}
We introduced \textbf{GeoTRAM}, a geometry-aware trajectory state modeling framework for few-shot action recognition. GeoTRAM constructs  relatively reliable 2.5D trajectories and models ordered intra-trajectory state evolution and state-conditioned local relations, while preserving complementary semantic, depth, and motion cues in dedicated subspaces before joint spatio-temporal reasoning. Dual-CLS-guided trajectory reweighting reduces action-irrelevant trajectories and guides selection, while complementary frame-level prototype and trajectory-native instance matching exploit aggregated frame-level evidence and fine-grained trajectory structure for few-shot classification. Experiments on five FSAR benchmarks show strong performance on motion-centric SSv2 datasets, competitive results on appearance-driven benchmarks, and a favorable accuracy--efficiency trade-off, supporting structured geometric trajectory-state modeling for FSAR.
\FloatBarrier

{
    \small
    \bibliographystyle{ieeenat_fullname}
    \bibliography{main}
}

\maketitlesupplementary
\appendix

This document provides supplementary material for the main manuscript and is organized as follows. Section~\ref{supp:quality-filtered-sampling} details the quality-filtered spectral point-sampling procedure. Section~\ref{supp:more_ablations} presents additional analyses, ablations, and visualizations. Section~\ref{supp:experiments} provides further implementation and experimental details.

\section{Quality-Filtered Spectral Point Sampling}
\label{supp:quality-filtered-sampling}

This section provides the implementation details of our quality-filtered
spectral point-sampling method. Given DINOv2~\cite{oquab2024dinov2}
patch features, we perform per-frame spectral partitioning, associate
the resulting clusters across time through a video-level prototype bank,
apply temporal and spatial quality filtering, select a peak frame for
each retained cluster, and allocate the point budget for seed extraction.
These steps together favor more semantically coherent, temporally stable, and spatially well-structured regions while reducing point initialization on unstable, fragmented, or isolated patches. The resulting point seeds are then used to initialize point tracking.

\subsection{Spectral Semantic Partitioning}
\label{supp:spectral-partitioning}

The goal of this stage is to partition each sampled frame into coherent
semantic regions without imposing a fixed number of clusters across
frames. Given a video, we uniformly select \(T\) frames and extract dense
DINOv2 features
\[
\mathcal{F}^{\mathrm{RGB}}
=
\{\mathbf{f}_{t,u}\}
\in
\mathbb{R}^{T\times H_p\times W_p\times C},
\]
where \(t\in\{1,\ldots,T\}\), \(u\in\{1,\ldots,N\}\) indexes a patch on
the \(H_p\times W_p\) grid, \(N=H_pW_p\), and \(C\) is the feature
dimension. Let \(\bar{\mathbf{f}}_{t,u}\) denote the
\(\ell_2\)-normalized descriptor of patch \(u\) in frame \(t\).

For each frame, we construct a symmetric
\(K_{\mathrm{nn}}\)-nearest-neighbor graph in feature space. An
undirected edge \((u,v)\) is retained if either patch is among the
\(K_{\mathrm{nn}}\) nearest neighbors of the other, with affinity
\begin{equation}
    w_{t,uv}
    =
    \exp\!\left[
        -\alpha
        \left(
            1-\bar{\mathbf{f}}_{t,u}^{\mathsf T}
            \bar{\mathbf{f}}_{t,v}
        \right)
    \right],
    \label{eq:spectral-affinity}
\end{equation}
where \(w_{t,uv}=w_{t,vu}\). We then perform spectral clustering using
the symmetric normalized graph Laplacian.

To adapt the partition to frame content, the number of clusters is
selected independently for each frame using the eigengap criterion.
Given the \(K_{\max}+1\) smallest eigenvalues
\(\lambda_{t,1}\leq\cdots\leq\lambda_{t,K_{\max}+1}\), we set
\begin{equation}
    K_t
    =
    \operatorname{clip}\left(
        \underset{1\leq k\leq K_{\max}}{\arg\max}
        \left(
            \lambda_{t,k+1}-\lambda_{t,k}
        \right),
        2,
        K_{\max}
    \right).
    \label{eq:eigengap}
\end{equation}
Let
\(\mathbf{U}_t\in\mathbb{R}^{N\times K_t}\) contain the eigenvectors
associated with the \(K_t\) smallest eigenvalues. K-means is applied to
its \(N\) row-wise spectral embeddings to obtain the per-frame patch
partition. Each resulting region is represented by the mean of its
normalized DINOv2 patch descriptors. In this way, the semantic
partition adapts its number of regions to the content of each frame
rather than using a fixed cluster count.

\subsection{Temporal Prototype Association}
\label{supp:temporal-association}

Because spectral clustering is performed independently in each frame,
its cluster indices are not temporally aligned. We therefore construct a
video-level prototype bank that associates semantically similar
per-frame regions and provides consistent labels across the sampled
frames.

The centroids from the first frame initialize the prototype bank. For a
centroid \(\mathbf{c}_{t,k}\) from a subsequent frame, we associate it with the most similar existing prototype; a new prototype is introduced only when it is sufficiently different from all existing prototypes:
\vspace{-1mm}
\begin{equation}
    \max_j
    \operatorname{cos}
    \left(
        \mathbf{c}_{t,k},
        \mathbf{p}_j
    \right)
    <
    \tau_{\mathrm{merge}},
    \label{eq:prototype-insertion}
    \vspace{-2mm}
\end{equation}
where \(\{\mathbf{p}_j\}\) denotes the current prototype bank. After the bank is constructed, every patch is assigned to its most similar video-level prototype by cosine similarity, yielding temporally aligned semantic labels
\(\mathcal{L}\in\mathbb{N}^{T\times H_p\times W_p}\).

\subsection{Quality Filtering and Budget Allocation}
\label{supp:quality-filtering}

Semantic consistency alone does not guarantee that a region provides
reliable tracking seeds. We therefore apply temporal and spatial quality
control before allocating the point budget. Temporal filtering is first
performed across the sampled frames. For each surviving cluster, we
then select a peak frame from its original temporally aligned support
and apply spatial filtering to the corresponding peak-frame mask.

\vspace{-6mm}
\paragraph{1) Temporal filtering.}
We first remove temporally unstable regions and suppress isolated label
changes.

\begin{itemize}
    \item \textbf{Temporal flicker.}
    For cluster \(j\), let \(q_{t,j}\) indicate its presence in frame
    \(t\). We discard the cluster if
    \[
        \sum_{t=1}^{T-1}
        |q_{t+1,j}-q_{t,j}|
        >
        \tau_{\mathrm{flicker}}.
    \]

    \item \textbf{Isolated label flips.}
    A solo flip occurs when the semantic label at the same patch-grid location changes between two consecutive sampled frames while none of its four spatial neighbors changes. Patch-grid locations accumulating more than \(\tau_{\mathrm{solo}}\)
    such events are excluded from the corresponding peak-frame masks.
    
\end{itemize}

\vspace{-5mm}
\paragraph{2) Peak-frame initialization.}
Each surviving video-level cluster is initialized only in the frame
where its semantic support is largest:
\vspace{-4mm}
\begin{equation}
    t_j^\star
    =
    \underset{t}{\arg\max}\;
    \sum_{u=1}^{N}
    \mathbb{I}[\ell_{t,u}=j].
    \label{eq:peak-frame}
\end{equation}
This avoids redundant initialization across time and places tracking seeds where the corresponding region has its largest spatial support.

\vspace{-4mm}
\paragraph{3) Spatial filtering.}
Spatial quality control is then applied to the selected peak-frame mask.

\begin{itemize}
    \item \textbf{Small components and isolated islands.}
    We reject clusters with fewer than \(n_{\min}\) patches in their
    selected peak frame. Using four-connectivity, we further reject a
    mask whose largest component contains fewer than two patches and
    remove singleton components whose minimum Euclidean patch-grid
    distance to the largest component exceeds \(\tau_{\mathrm{island}}\).

    \item \textbf{Spatial compactness.}
    For cluster support \(\Omega_{t,j}\), we measure its occupancy
    relative to its bounding box:
    \begin{equation}
        \rho_{t,j}
        =
        \frac{
            |\Omega_{t,j}|
        }{
            \operatorname{area}
            \left(
                \operatorname{BBox}(\Omega_{t,j})
            \right)
        }.
        \label{eq:compactness}
    \end{equation}
    A cluster is marked as compact if
    \(\rho_{t,j}\geq\tau_{\mathrm{compact}}\), and as scattered
    otherwise.
\end{itemize}

\vspace{-3mm}
\paragraph{4) Reliability-aware budget allocation.}
We guarantee each retained cluster a base allocation and assign the
remaining point budget preferentially to spatially compact regions.
Let \(B\) denote the target video-level budget, \(b\) the base
allocation, and \(n_g\) and \(n_s\) the numbers of compact and scattered
clusters, respectively. When
\(b(n_g+n_s)\leq B\), scattered clusters receive \(m_s=b\) points,
whereas compact clusters receive
\begin{equation}
    m_g
    =
    b+
    \left\lfloor
        \frac{
            B-b(n_g+n_s)
        }{
            \max(n_g,1)
        }
    \right\rfloor.
    \label{eq:point-budget}
\end{equation}
Thus, the remaining capacity is assigned to the more
spatially coherent regions. If the base allocation exceeds \(B\), both
allocations are reduced by a common factor while retaining at least one
point per cluster. Any residual budget after integer allocation is
distributed among the retained clusters, ensuring exactly \(B\)
tracking seeds.

\subsection{Seed Extraction}
\label{supp:seed-extraction}
For each retained cluster, we upsample the filtered peak-frame mask to
image resolution using nearest-neighbor interpolation and perform
coordinate K-means over the pixels inside the mask. The nearest valid
mask pixel to each centroid is selected as a tracking seed, ensuring
spatial coverage while keeping all queries within the semantic region.
If fewer valid mask pixels than allocated seeds are available, sampling
with replacement is used to meet the assigned allocation. We uniformly sample \(T=32\) frames for the
quality-filtered seed-initialization stage and extract DINOv2~\cite{oquab2024dinov2} ViT-B/14
patch tokens from \(224\times224\) inputs
(\(H_p=W_p=16\)). The denser temporal sampling is used only for offline
point initialization, providing broader temporal coverage for prototype
association, temporal quality filtering, and peak-frame selection; the
downstream GeoTRAM representation uses \(T=8\) frames per video. We use \(K_{\mathrm{nn}}=20\), \(\alpha=5\), \(K_{\max}=12\), and
\(\tau_{\mathrm{merge}}=0.2\). We set
\(\tau_{\mathrm{flicker}}=5\),
\(\tau_{\mathrm{solo}}=5\),
\(n_{\min}=3\),
\(\tau_{\mathrm{island}}=3\) patch units, and
\(\tau_{\mathrm{compact}}=0.3\).
The target point budget is \(B=256\) seeds per video with base allocation
\(b=16\). Each seed's initialization frame is mapped from the sampled clustering
frames to the full-video timeline, and the resulting queries are passed
to CoTracker~\cite{karaev2024cotracker} for bidirectional tracking.

\begin{figure}[!h]
    \centering
    \includegraphics[width=0.99\linewidth]{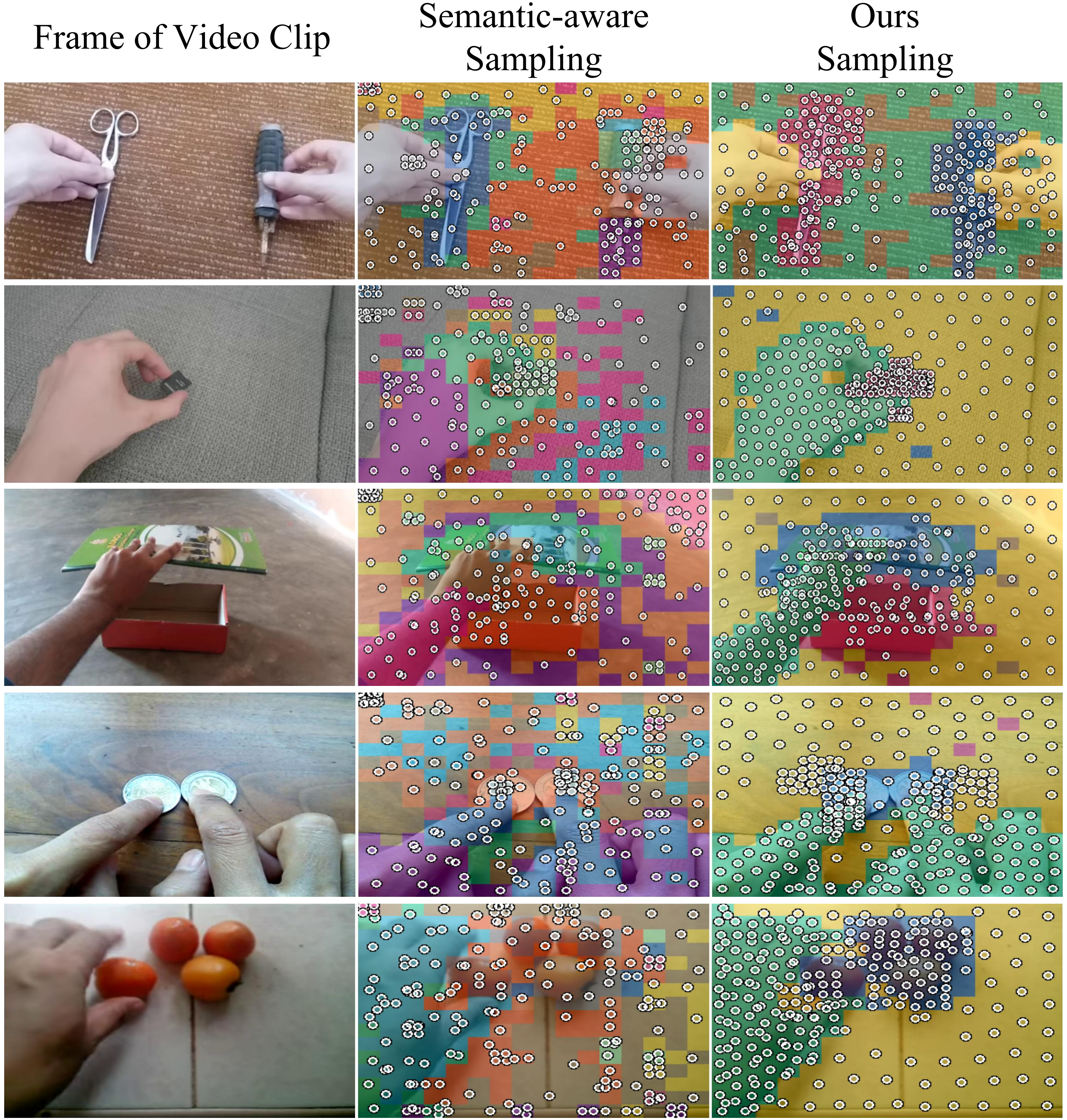}
    \caption{Qualitative comparison between Trokens' semantic-aware point sampling and our quality-filtered spectral point sampling strategies. For each video, the semantic clusters and sampled point locations are visualized on the frame. Semantic-aware sampling often produces fragmented or redundant samples across multiple semantic regions, whereas our quality-filtered spectral sampling suppresses isolated and scattered regions and allocates more seeds to spatially coherent hand--object interaction regions.}
    \vspace{-3mm}
    \label{fig:point_sampling_comparison}
\end{figure}

\begin{figure*}[!t]
    \centering
    \includegraphics[width=0.90\linewidth]{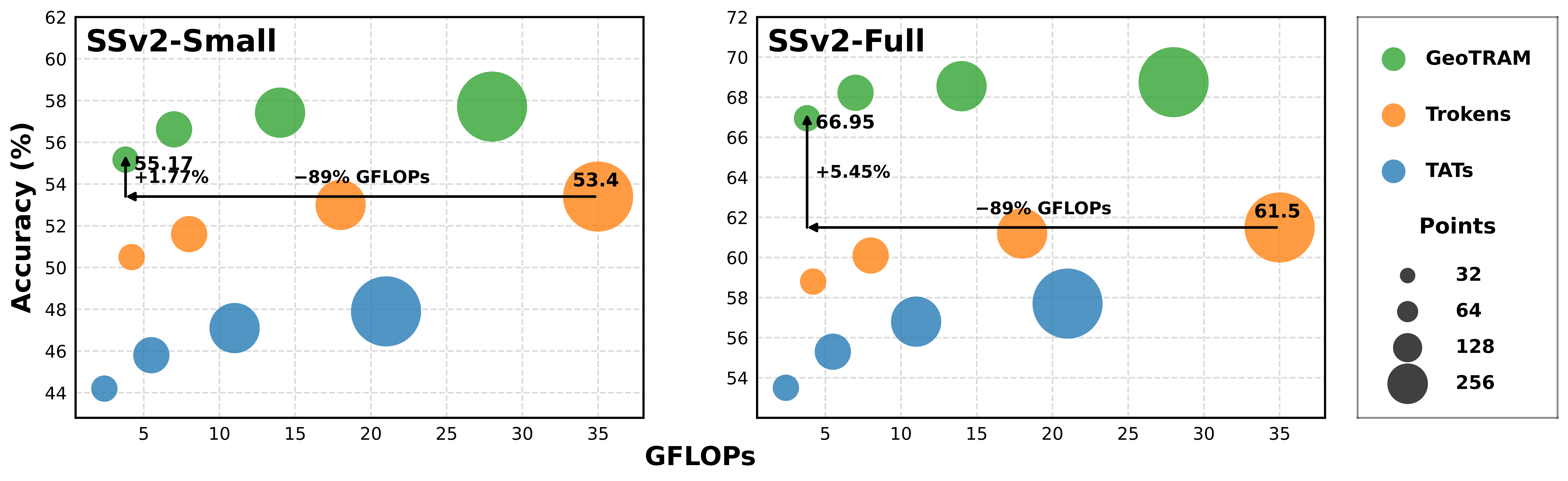}
    \vspace{-2mm}
    \caption{Accuracy--efficiency comparison on SSv2-Small and SSv2-Full under the 5-way 1-shot setting with different trajectory point counts. Bubble size denotes the number of tracked points. GeoTRAM consistently outperforms Trokens and TATs across the evaluated point counts, and its 32-point configuration already surpasses 256-point Trokens while reducing GFLOPs by 89\%.}
    \vspace{-3mm}
    \label{fig:flops_accuracy_points}
\end{figure*}

\section{Additional Analyses and Ablations}
\label{supp:more_ablations}

\subsection{Point-Sampling Comparison with Baselines.}
\label{supp:point_sampling_comparison}
Figure~\ref{fig:point_sampling_comparison} qualitatively compares the semantic-aware sampling of Trokens~\cite{kumar2025trokens} with our quality-filtered spectral sampling strategy. For context, uniform sampling distributes points evenly over the entire frame and can therefore place many tracking seeds on static or action-irrelevant background regions. Semantic-aware sampling provides semantically informed initialization, but can still produce fragmented or redundant samples across multiple regions and retain seeds in spatially scattered areas. Our method instead applies explicit temporal and spatial quality filtering and allocates more seeds to spatially coherent hand--object interaction regions. As a result, the selected tracking points provide a more reliable foundation for constructing the trajectory evidence used throughout GeoTRAM, benefiting subsequent geometric motion modeling, multi-cue representation, and few-shot matching.

\begin{figure}[!t]
    \centering
    \includegraphics[width=0.99\linewidth]{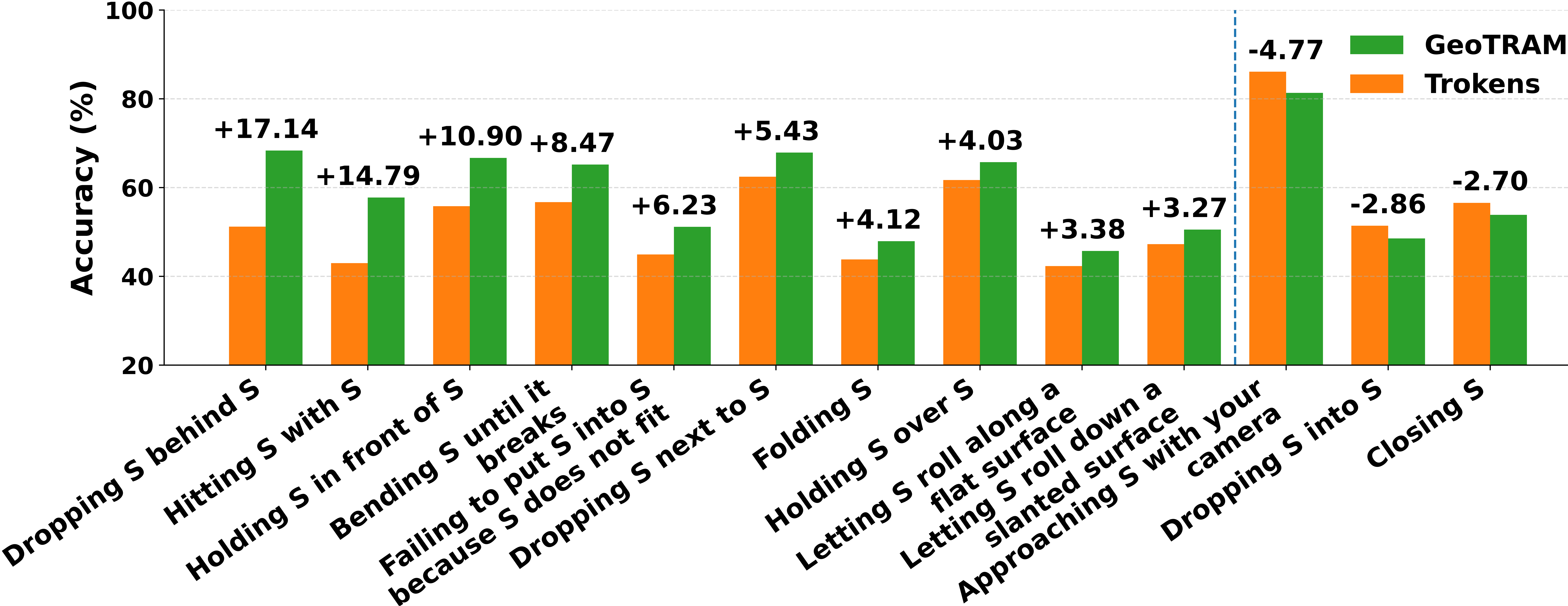}
    \vspace{-2mm}
    \caption{Class-wise accuracy comparison between Trokens and GeoTRAM on SSv2-Small. The top-10 gains (left) and three largest drops (right) are shown. Many of the largest gains occur for actions involving structured object motion and relative spatial relations.}
    \label{fig:classwise}
    \vspace{-5mm}
\end{figure}

\subsection{Effect of Trajectory Point Count.}
\label{supp:pointcount_effect}

Figure~\ref{fig:flops_accuracy_points} compares the accuracy--efficiency trade-off of GeoTRAM, Trokens~\cite{kumar2025trokens}, and TATs~\cite{kumar2024tats} under the 5-way 1-shot setting with different trajectory point counts. Across both SSv2-Small and SSv2-Full, GeoTRAM achieves the highest accuracy at every evaluated point count while maintaining a favorable accuracy--efficiency trade-off. Notably, with only 32 points, GeoTRAM achieves 55.17\% and 66.95\% accuracy on SSv2-Small and SSv2-Full, respectively, outperforming 256-point Trokens by 1.77 and 5.45\% points while reducing GFLOPs from 35 to 3.8, corresponding to an 89\% reduction. These results show that GeoTRAM retains strong discriminative performance even under a substantially reduced trajectory budget, producing a markedly better accuracy--efficiency trade-off.

\subsection{Class-wise analysis.}
\label{supp:classwise_analysis}
Figure~\ref{fig:classwise} shows that many of the largest gains occur for actions characterized by structured object motion and relative spatial relations, such as ``Dropping something behind something'' and ``Holding something in front of something.'' Smaller reductions are observed for a few classes, including ``Approaching something with your camera'' and ``Closing something,'' ,indicating that the improvements are not uniform across all action categories. This pattern is consistent with GeoTRAM's design goal of modeling ordered motion dynamics and relative geometry for fine-grained action discrimination.

\begin{table}[!t]
    \centering
    \scriptsize
    \caption{Stability of GeoTRAM across three independently trained runs
    with different random seeds. Accuracy (\%) is reported as mean $\pm$
    standard deviation across runs.}
    \label{tab:variance}
    \begin{tabular}{lccc}
    \toprule
    \textbf{Datasets} & \textbf{5-way 1-shot} & \textbf{5-way 3-shot} & \textbf{5-way 5-shot} \\
    \midrule
    SSv2-Full~\cite{goyal2017something}  & 68.78 $\pm$ 0.23 & 79.80 $\pm$ 0.20 & 82.41 $\pm$ 0.18 \\
    SSv2-Small~\cite{goyal2017something} & 57.55 $\pm$ 0.31 & 70.41 $\pm$ 0.27 & 74.14 $\pm$ 0.27 \\
    Kinetics~\cite{carreira2017quo}      & 83.15 $\pm$ 0.22 & 90.23 $\pm$ 0.18 & 91.82 $\pm$ 0.16 \\
    UCF-101~\cite{ucf101}                & 93.40 $\pm$ 0.30 & 97.55 $\pm$ 0.25 & 98.11 $\pm$ 0.22 \\
    HMDB-51~\cite{hmdb51}                & 71.13 $\pm$ 0.29 & 81.15 $\pm$ 0.25 & 83.91 $\pm$ 0.20 \\
    \bottomrule
    \end{tabular}
    \vspace{-3mm}
\end{table}

\subsection{Variance over multiple training runs.}
\label{supp:variance}

Table~\ref{tab:variance} reports the performance of GeoTRAM across three independently trained runs with different random seeds. The results show relatively low running variation across all five dataset configurations and shot settings, with standard deviations ranging from 0.16 to 0.31\%. These results indicate that the performance of GeoTRAM is stable across different training runs and is not too sensitive to a particular random seed.

\section{Experiments}
\label{supp:experiments}

\subsection{Additional training and evaluation details}
\label{sec:supp-impl}

For consistency with prior trajectory-based FSAR methods~\cite{kumar2024tats,kumar2025trokens},
we largely follow their training protocol for optimization and episodic-training hyperparameters. We optimize all trainable modules with AdamW for $50$ epochs on a single NVIDIA A100-40GB GPU using mixed precision. The learning rate is linearly warmed up from $1{\times}10^{-6}$ to the base learning rate $1{\times}10^{-4}$ over the first $5$ epochs and is then decayed by a factor of $0.1$ at epochs $20$ and $40$. The weight decay is $0.5$. Random spatial cropping is the only data augmentation used during training.

For the default configuration, we uniformly sample $T=8$ frames per video and resize/crop them to $224\times224$, followed by ImageNet mean and standard-deviation normalization. We use $M=256$ tracking seeds per video. The seeds are mapped to the original video timeline and tracked offline with CoTracker~\cite{karaev2024cotracker}; the resulting trajectories are reused across episodes. Depth maps are estimated online using the frozen DPT-based depth-estimation head~\cite{ranftl2021vision} applied to intermediate DINOv2~\cite{oquab2024dinov2} features. A $P\times P$ local depth patch with $P=14$ is sampled around each tracked point. DINOv2 ViT-B/14~\cite{oquab2024dinov2}, the DPT-based depth head, and CoTracker~\cite{karaev2024cotracker} remain frozen.

StateMotion uses a two-layer temporal 1D-convolutional encoder. Before multi-cue fusion, each of the four cue branches---semantic, intra-motion, inter-motion, and depth---is independently refined by a temporal self-attention block with $4$ heads, an FFN with hidden size $512$, and sinusoidal temporal positional encodings before concatenation. The fused tokens are processed by a single divided space--time Transformer block following the design of TATs~\cite{kumar2024tats}. StateMotion and RelMotion use an internal hidden size of $128$ and dropout $0.1$, and RelMotion uses $k=8$ nearest neighbors. Dual-CLS fusion uses $\beta=0.5$. The dual branch classifier combines the frame-level prototype branch and the trajectory-level instance branch, with top-$\rho$ trajectory selection ($\rho=0.5$) applied only to the trajectory-level branch. We use $(C_{\mathrm{sem}}, C_{\mathrm{motion}}, C_{\mathrm{dep}}, C)= (384,176,32,768)$ for GeoTRAM and $(240,144,32,560)$ for the compact GeoTRAM-S variant, where $C=C_{\mathrm{sem}}+C_{\mathrm{dep}}+2C_{\mathrm{motion}}$.

Following standard FSAR protocols, both training and evaluation use 5-way episodic sampling. Each training episode contains $K$ support videos and $6$ query videos per class, corresponding to $5(K+6)$ videos in total. At evaluation, each episode contains $K$ support videos and one query video per class, corresponding to $5(K+1)$ videos. We report 5-way $K$-shot accuracy averaged over $10{,}000$ test episodes. Unless otherwise stated, fixed random seeds are used for reproducibility, and the last-epoch checkpoint is used for evaluation.

\subsection{Extended Comparison with Prior Work}
\label{sec:supp-tables}

Due to the page limit, the main paper reports comparisons with a representative subset of prior FSAR methods. Tables~\ref{supp_table1} and~\ref{supp_table2} extend the comparison to a broader set of representative FSAR approaches following standard benchmark splits and episodic evaluation protocols. We include both classical and recent methods to provide broader context, while TATs~\cite{kumar2024tats} and Trokens~\cite{kumar2025trokens} are the most directly related trajectory-based competitors. The extended results remain consistent with the conclusions of the main paper.

\begin{table*}[!t]
\centering
\renewcommand{\arraystretch}{0.80}

\caption{A more comprehensive comparison of few-shot action
recognition accuracy (\%) under 1--5 shot settings on
SSv2-Full and Kinetics. Best results are in \textbf{bold},
second-best results are \underline{underlined}.}
\label{supp_table1}

\resizebox{0.9\textwidth}{!}{
\begin{tabular}{lcccccccccccc}

\toprule
\textbf{Method} & \textbf{Venue} &
\multicolumn{5}{c}{\textbf{SSv2-Full}} &
\multicolumn{5}{c}{\textbf{Kinetics}} \\
\cmidrule(lr){3-7}\cmidrule(lr){8-12}
& & 1-shot & 2-shot & 3-shot & 4-shot & 5-shot
  & 1-shot & 2-shot & 3-shot & 4-shot & 5-shot \\
\midrule
OTAM~\cite{cao2020otam} & CVPR'20 & 42.8 & 49.1 & 51.5 & 52.0 & 52.3 & 72.2 & 75.9 & 78.7 & 81.9 & 84.2 \\
TRX~\cite{perrett2021trx} & CVPR'21 & 42.0 & 53.1 & 57.6 & 61.1 & 64.6 & 63.6 & 76.2 & 81.8 & 83.4 & 85.2 \\
STRM~\cite{thatipelli2022strm} & CVPR'22 & 43.1 & 53.3 & 59.1 & 61.7 & 68.1 & 62.9 & 76.4 & 81.1 & 83.8 & 86.7 \\
MTFAN~\cite{wu2022motion} & CVPR'22 & 45.7 & -- & -- & -- & 60.4 & 74.6 & -- & -- & -- & 87.4 \\
HyRSM~\cite{wang2022hyrsm} & CVPR'22 & 54.3 & 62.2 & 65.1 & 67.9 & 69.0 & 73.7 & 80.0 & 83.5 & 84.6 & 86.1 \\
HCL~\cite{zheng2022few} & ECCV'22 & 47.3 & 54.5 & 59.0 & 62.4 & 64.9 & 73.7 & 79.1 & 82.4 & 84.0 & 85.8 \\
Nguyen et al.~\cite{nguyen2022inductive} & ECCV'22 & 43.8 & -- & -- & -- & 61.1 & 74.3 & -- & -- & -- & 87.4 \\
Huang et al.~\cite{huang2022compound} & ECCV'22 & 49.3 & -- & -- & -- & 66.7 & 73.3 & -- & -- & -- & 86.4 \\
MoLo~\cite{wang2023molo} & CVPR'23 & 56.6 & 62.3 & 67.0 & 68.5 & 70.6 & 74.0 & 80.4 & 83.7 & 84.7 & 85.6 \\
SloshNet\cite{xing2023SloshNet} & AAAI'23 & 46.5 & -- & -- & -- & 68.3 & 70.4 & -- & -- & -- & 87.0 \\
GgHM~\cite{xing2023gghm} & ICCV'23 & 54.5 & -- & -- & -- & 69.2 & 74.9 & -- & -- & -- & 87.4 \\
RFPL~\cite{xia2023rfpl} & ICCV'23 & 47.0 & 54.6 & 58.3 & 60.3 & 61.0 & 74.6 & 80.0 & 82.1 & 84.1 & 86.8 \\
CCLN~\cite{wang2024ccln} & TIP'24 & 46.0 & -- & -- & -- & 61.3 & 75.8 & 82.1 & 85.0 & 86.1 & 87.5 \\
HyRSM++~\cite{wang2024hyrsm++} & PR'24 & 55.0 & 63.5 & 66.0 & 68.8 & 69.8 & 74.0 & 80.8 & 83.9 & 85.3 & 86.4 \\
TATs~\cite{kumar2024tats} & ECCV'24 & 57.7 & 67.1 & 70.0 & 70.6 & 74.6 & 81.9 & 86.5 & \underline{89.9} & 90.6 & 91.1 \\
TEAM~\cite{lee2025team} & CVPR'25 & -- & -- & -- & -- & -- & \textbf{83.3} & -- & -- & -- & \textbf{92.9} \\
Trokens~\cite{kumar2025trokens} & ICCV'25 & \underline{61.5} & \underline{69.9} & \underline{73.8} & \underline{75.9} & \underline{76.7} & 82.9 & \textbf{87.7} & \underline{89.9} & \underline{90.8} & 91.2 \\

\midrule
\textbf{GeoTRAM} & WACV'27 & \textbf{68.9} & \textbf{76.5} & \textbf{79.9} & \textbf{81.9} & \textbf{82.3} & \underline{83.1} & \underline{87.2} & \textbf{90.2} & \textbf{91.0} & \underline{91.8} \\

\bottomrule
\end{tabular}}
\label{supp_table1}
\vspace{6mm}

\caption{A more comprehensive comparison of few-shot action
recognition accuracy (\%) under 1-, 3-, and 5-shot settings
on SSv2-Small, UCF-101, and HMDB-51. Best results are
in \textbf{bold}, second-best results are
\underline{underlined}.}
\label{supp_table2}

\resizebox{0.9\textwidth}{!}{
\begin{tabular}{lcccccccccc}
\toprule
\textbf{Method} & \textbf{Venue} &
\multicolumn{3}{c}{\textbf{SSv2-Small}} &
\multicolumn{3}{c}{\textbf{UCF-101}} &
\multicolumn{3}{c}{\textbf{HMDB-51}} \\
\cmidrule(lr){3-5}\cmidrule(lr){6-8}\cmidrule(lr){9-11}
& & 1-shot & 3-shot & 5-shot
  & 1-shot & 3-shot & 5-shot
  & 1-shot & 3-shot & 5-shot \\
\midrule
OTAM~\cite{cao2020otam} & CVPR'20 & 36.4 & 45.9 & 48.0 & 79.9 & 87.0 & 88.9 & 54.5 & 65.7 & 68.0 \\
TRX~\cite{perrett2021trx} & CVPR'21 & 36.0 & 51.9 & 56.7 & 78.2 & 92.4 & 96.1 & 53.1 & 66.8 & 75.6 \\
STRM~\cite{thatipelli2022strm} & CVPR'22 & 37.1 & 49.2 & 55.3 & 80.5 & 92.7 & 96.9 & 52.3 & 67.4 & 77.3 \\
MTFAN~\cite{wu2022motion} & CVPR'22 & -- & -- & -- & 84.8 & -- & 95.1 & -- & -- & -- \\
HyRSM~\cite{wang2022hyrsm} & CVPR'22 & 40.6 & 52.3 & 56.1 & 83.9 & 93.0 & 94.7 & 60.3 & 71.7 & 76.0 \\
HCL~\cite{zheng2022few} & ECCV'22 & 38.7 & 49.1 & 55.4 & 82.5 & 91.0 & 93.9 & 59.1 & 71.2 & 76.3 \\
Nguyen et al.~\cite{nguyen2022inductive} & ECCV'22 & -- & -- & -- & -- & -- & -- & 59.6 & -- & 76.9 \\
Huang et al.~\cite{huang2022compound} & ECCV'22 & 38.9 & -- & 61.6 & 71.4 & -- & 91.0 & 60.1 & -- & 77.0 \\
MoLo~\cite{wang2023molo} & CVPR'23 & 42.7 & 52.9 & 56.4 & 86.0 & 93.5 & 95.5 & 60.8 & 72.0 & 77.4 \\
GgHM~\cite{xing2023gghm} & ICCV'23 & -- & -- & -- & 85.2 & -- & 96.3 & 61.2 & -- & 76.9 \\
RFPL~\cite{xia2023rfpl} & ICCV'23 & -- & -- & -- & 82.5 & 94.1 & 96.3 & -- & -- & -- \\
CCLN~\cite{wang2024ccln} & TIP'24 & -- & -- & -- & 86.9 & 94.2 & 96.1 & 65.1 & 76.2 & 78.8 \\
HyRSM++~\cite{wang2024hyrsm++} & PR'24 & 42.8 & 52.4 & 58.0 & 85.8 & 93.5 & 95.9 & 61.5 & 72.7 & 76.4 \\
TATs~\cite{kumar2024tats} & ECCV'24 & 47.9 & 60.0 & 64.4 & 92.0 & 96.8 & 95.5 & 60.0 & 71.8 & 77.0 \\
TEAM~\cite{lee2025team} & CVPR'25 & 47.2 & -- & 63.1 & \textbf{94.5} & -- & \textbf{98.8} & \underline{70.9} & -- & \textbf{85.5} \\
Trokens~\cite{kumar2025trokens} & ICCV'25 & \underline{53.4} & \underline{65.3} & \underline{68.9} & \underline{94.0} & \underline{97.3} & 97.9 & 69.8 & \underline{80.0} & 82.3 \\

\midrule
\textbf{GeoTRAM} & WACV'27 & \textbf{57.7} & \textbf{70.4} & \textbf{74.1} & 93.5  & \textbf{97.5} & \underline{98.2} & \textbf{71.2} & \textbf{81.0} & \underline{83.9} \\
\bottomrule
\end{tabular}}
\label{supp_table2}
\end{table*}

\end{document}